# A Multi-disciplinary Ensemble Algorithm for Clustering Heterogeneous Datasets

Bryar A. Hassan[1,3], Tarik A. Rashid[2]

[1]Kurdistan Institution for Strategic Studies and Scientific Research, Sulaimani, Iraq

[2]Computer Science and Engineering Department, University of Kurdistan Hewler, Iraq

[3]Department of Computer Networks, Technical College of Informatics, Sulaimani Polytechnic University, Sulaimani, Iraq

Email: bryar.hassan@kissr.edu.krd

## Abstract

Clustering is a commonly used method for exploring and analysing data where the primary objective is to categorise observations into similar clusters. In recent decades, several algorithms and methods have been developed for analysing clustered data. We notice that most of these techniques deterministically define a cluster based on the value of the attributes, distance, and density of homogenous and single-featured datasets. However, these definitions are not successful in adding clear semantic meaning to the clusters produced. Evolutionary operators and statistical and multi-disciplinary techniques may help in generating meaningful clusters. Based on this premise, we propose a new evolutionary clustering algorithm (ECA*) based on social class ranking and meta-heuristic algorithms for stochastically analysing heterogeneous and multiple-featured datasets. The ECA* is integrated with recombinational evolutionary operators, Levy flight optimisation, and some statistical techniques, such as quartiles and percentiles, as well as the Euclidean distance of the K-means algorithm. Experiments are conducted to evaluate the ECA* against five conventional approaches: K-means (KM), K-means++ (KM++), expectation maximisation (EM), learning vector quantisation (LVQ), and the genetic algorithm for clustering++ (GENCLUST++). That the end, 32 heterogeneous and multiple-featured datasets are used to examine their performance using internal and external and basic statistical performance clustering measures and to measure how their performance is sensitive to five features of these datasets (cluster overlap, the number of clusters, cluster dimensionality, the cluster structure, and the cluster shape) in the form of an operational framework. The results indicate that the ECA* surpasses its counterpart techniques in terms of the ability to find the right clusters. Significantly, compared to its counterpart techniques, the ECA* is less sensitive to the five properties of the datasets mentioned above. Thus, the order of overall performance of these algorithms, from best performing to worst performing, is the ECA*, EM, KM++, KM, LVQ, and the GENCLUST++. Meanwhile, the overall performance rank of the ECA* is 1.1 (where the rank of 1 represents the best performing algorithm and the rank of 6 refers to the worst-performing algorithm) for 32 datasets based on the five dataset features mentioned above.

## 1. Introduction

Clustering groups the observations of a dataset in a way that similar observations are put together in a cluster, and dissimilar observations are positioned in different clusters.

Clustering algorithms have a wide range of applications in nearly all fields, including ontology learning, medical imaging, gene analysis, and market analysis. In the past, several clustering techniques have been proposed, but each of these algorithms was mostly dedicated to a specific type of problems [1]. For instance, [2] concluded that K-means works poorly for datasets that have a significant number of clusters, cluster size, or imbalance among clusters. Not every clustering algorithm needs to perform well in all or almost all cohorts of datasets and real-world applications. Instead, it is necessary to show how sensitive an algorithm is to different types of benchmarking and real-world problems. Nevertheless, almost all clustering techniques have some common limitations: (1) Finding the right number of clusters is a challenging task [3]. (2) The sensitivity of current clustering algorithms to a random selection of cluster centroids is another constraint of clustering techniques since selecting bad cluster centroids can easily lead to inadequate clustering solutions [4]. (3) Since almost every clustering technique engages in the process of hill-climbing towards its objective function, they can easily become stuck in local optima, which may lead to poor clustering results [5]. (4) Data are not free from outliers and noise. Nonetheless, we assume that clusters have a similar spread and a roughly equal density. As a consequence, noise and outliers may misguide clustering techniques to produce an excellent clustering result. (5) There are a minimal number of studies that show the sensitivity of clustering algorithms to cohorts of datasets and real-world applications. (6) Most techniques and algorithms define a cluster in terms of the value of attributes, density, and distance. Nevertheless, these definitions might not give an exact clustering result [6]. (7) Moreover, most prior algorithms are deterministic [1], and their clustering results are entirely related to their initial states and inputs. The given starting conditions and initial input parameters may directly affect the generation of output, which may achieve a balance between exploitation and exploration search spaces and ensure that clustering algorithms cannot easily trap in local and global optima accordingly [7].

Therefore, there is an enormous demand for developing a new clustering algorithm that is free from the shortcomings of current clustering techniques. Because we integrate several techniques in our proposed algorithm, we call it the ECA*. The techniques integrated into the ECA* are social class ranking, quartiles, percentiles and percentile ranking, evolutionary algorithm operators, the involvement of randomness, and the Euclidean distance in the K-means algorithm. Moreover, the evolutionary algorithm operators are crossover, mutation, a crossover-mutation recombination strategy called mut-over in the backtracking search optimisation algorithm (BSA), and the random walk in the Levy flight optimisation (LFO).

One crucial element of the ECA* is the use of two evolutionary algorithm operators: (i) The BSA is a popular meta-heuristic algorithm. It consists of the following elements: Initialisation, Selection I, Mutation, Crossover, Selection II, and Fitness Evaluation [7]. The essential BSA operators are mutation and crossover since these operators are in the BSA as a recombination-based strategy. Crossover and mutation operators are also used in genetic algorithms and differential evolutionary algorithms. Optimisation operators are used in this study for two reasons [7, 8]: First, the BSA is a popular meta-heuristic algorithm that uses the DE crossover and mutation operators as a recombination-based strategy. This recombination helps the BSA to outperform its counterpart algorithms. Additionally, the BSA strikes a balance between local and global searches to avoid becoming trapped in local and global optima, and it provides robustness and stability to our proposed technique. Second, LFO is a basic technique that uses several particles in each generation. From a better-known spot, the algorithm can create a whole new generation at distances randomly distributed by Levy flights. Then, the new generation is tested to choose the most promising solution. The process is repeated until the stop conditions have been met. LFO implementation is straightforward [9]. Only the best solution is selected. The pseudo-code algorithm based on the LFO concept is detailed in [9].

Another significant element of this algorithm is the utilisation of social class ranking. Each cluster is considered a social class rank and is represented by a percentile rank in the dataset. Every society is classified based on economic and social statuses. Social class is the rank of people within a society based on their characteristics, such as their resources, authority, and power [10]. Different societies in different countries have different numbers of social class ranks. Many sociologists prefer five classes: elite and upper class, upper-middle class, lower-middle class, working-class, and poor class. This real-life ranking classification is practised in the proposed algorithm. Primarily, clustering deals with distributing a set and observations into clusters based on their attributes. Since clustering is an unsupervised learning technique, the number of clusters or groups of data is previously unknown. This notion leads us to design the proposed algorithm based on an adaptive number of clusters. In the proposed algorithm, each data observation belongs to a cluster based on its percentile rank in the dataset. Table 1 represents an instance of a class name, cluster representation, and percentile rank boundary for each cluster in the proposed algorithm. This notion leads us to design the proposed algorithm based on an adaptive number of clusters.

Table 1: A sample representation of the social class ranking

| Class name | Cluster representation | Percentile rank |
| --- | --- | --- |
| Upper cluster (Elite) | Cluster1 | 80-100 |
| Middle Cluster (Upper) | Cluster2 | 60-80 |

| Middle cluster (Lower) | Cluster3 | 40-60 |
| Working-cluster | Cluster4 | 20-40 |
| Poor cluster | Cluster5 | 0-20 |

There are several motives behind our proposal of the ECA*. First, suggesting a dynamic clustering strategy to determine the right number of clusters is a remarkable achievement in clustering techniques. This strategy removes empty clusters, merges dynamic clusters that are less dense with their neighbouring clusters and merging two clusters based on their distance measures. Another motive is the desire to find the right cluster centroids for the clustering results. Outliers and noise are always part of data, and they sometimes mislead the clustering results. To prevent this deviation, quartile boundaries and evolutionary operators can be exploited. These methods are useful for adjusting clustering centroids towards their objective function and result in excellent clustering centroids accordingly. In addition, developing a new clustering algorithm that does not become trapped in local and global optima is an incentive. In turn, a boundary control mechanism can be defined as a very effective technique in gaining diversity and ensuring efficient searches for finding the best cluster centroids. It is also concluded that having statistical techniques and the involvement of randomness in clustering techniques can help to generate meaningful clusters in a set of numerical data [6]. Based on this premise, it is necessary to develop a stochastic clustering algorithm to seek a balance between exploration and exploitation search spaces and to produce an excellent clustering result accordingly. Finally, defining an approach for clustering multiple-features and different types of problems is another rationale behind our introduction of the ECA*.

Insufficient research has been conducted to study the behaviour of recent clustering algorithms against various types of dataset problems. As an exception, this research [2] introduced basic clustering benchmark datasets to determine how the K-means algorithm behaves under different cohorts and features of dataset problems. The dataset types used in the study are A, S, G2, DIM, Birch, and Unbalance. These datasets have different clustering dataset features, such as cluster overlap, cluster dimensionality, imbalance in cluster size, and the number of clusters.

Clustering can take two primary forms: (i) In the presence of cluster labels; a clustering technique is called supervised analysis. (ii) Unsupervised clustering techniques do not require cluster labels in datasets as a target pattern for the output. This form of clustering is the focus of our study. In addition to introducing these two primary forms of clustering techniques, a hybrid clustering technique that combines supervised and unsupervised techniques has been proposed to provide compelling prediction performance in the case of unbalanced datasets [11]. The critical contributions of the ECA* are as follows: At first glance, we integrate the notion of social class ranking, quartiles and percentiles, recombined operators of meta-heuristic algorithms, the randomness process, and the Euclidean distance in the K-means algorithm as combinational techniques of our introduced technique. In Section 3, we conduct an analysis to present the advantages of the above techniques in our proposed algorithm. One of the essential benefits of the ECA* is that it finds the right number of clusters. In almost all current algorithms, the number of clusters should be previously defined. In contrast to prior clustering algorithms, the most crucial contribution of the ECA* is its adaptivity in determining the number of clusters. The ECA* has an adaptive number of clusters since it is based on social class ranking. The motives behind using social class ranking are to provide an adaptive number of clusters in every clustering dataset problem. The social rankings initially set the number of clusters, and then, the empty clusters are removed. Additionally, clusters that are less dense will be merged with their neighbouring clusters. Furthermore, the selection of the right cluster centroids is one of the contributions of the ECA*. Three steps are used for computing cluster centroids as follows: (i) Initially, cluster centroids are calculated based on the mean quartile of the clusters. (ii) Historical cluster centroids are selected between the lower and upper quartiles to prevent outliers and noise from misleading the algorithm. (iii) Meta-heuristic algorithm operators are used to adjusting the cluster centroids to produce an excellent clustering result. Another advantage of the ECA* is the involvement of stochastic and randomness processes. The advantage of the stochastic process in the ECA* is that it strikes a balance between exploring the search space and utilising the learning process in the search space to hone in on local and global optima. Furthermore, the excellent performance of the ECA* is attributable to the use of meta-heuristic algorithm operators [8]. Meanwhile, the mutation strategy of our algorithm is improved in two ways [12]. First, the adaptive control parameter (F) is introduced by using Levy flight optimisation to balance the exploitation and exploration of the algorithm. Second, the cluster centroids learn knowledge from the historical cluster centroids (HI) to increase the learning ability of the cluster centroids and to find the best cluster centroids for clusters accordingly. Mut-over can also provide a mutation-crossover recombination strategy. Third, mut-over with the aid of F and HI can overcome the issue of global and local optima that may occur in other algorithms, such as K-means [13]. This recombined evolutionary operator also provides stability and robustness to our proposed algorithm [12]. Hence, these strategies correctly balance the relationship between global optima and local optima. Finally, another contribution of the ECA* is that it can handle scalable and multi-feature datasets with metric attributes, such as two-dimensional and multi-dimensional datasets with a different number of clusters, as well as overlapped, well-structured, unstructured, and dimensional clusters. The components of the ECA* are depicted in Figure 1 as a basic flowchart.

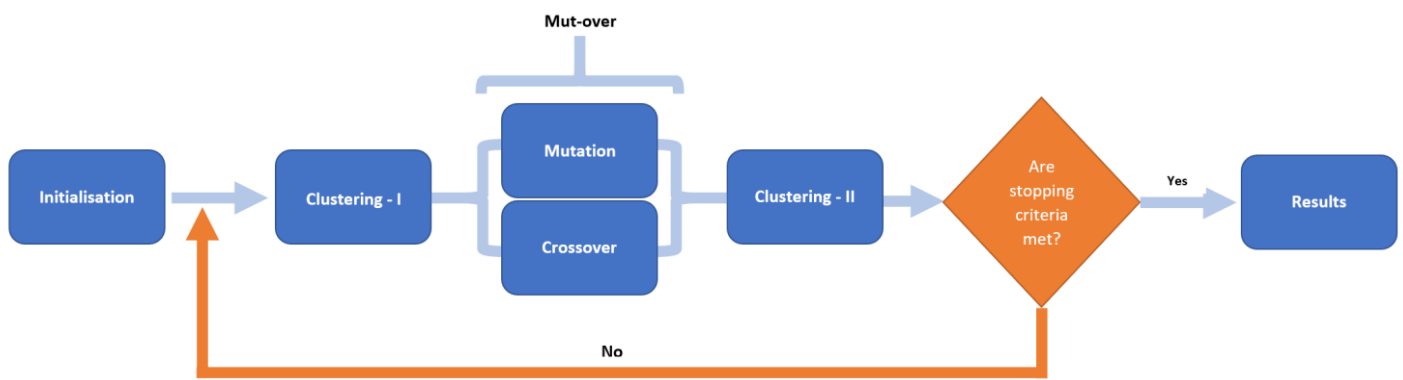

**Figure 1: The basic flowchart of ECA*.**

The rest of this study is organised as follows: Section 2 reviews the studies related to clustering techniques. In Section 3, our proposed algorithm is detailed. The experiments and result analysis are discussed in Section 4. The experimental results are presents in Section 5, followed by a proposed framework to present how the algorithms perform based on five dataset features. Finally, Section 6 draws concluding remarks and reflects on the strengths and weaknesses of the ECA* to explain how scholars can benefit from this work, followed by directions for future work.

## 2. Related work

Many clustering algorithms have been developed recently. K-means is one of the most popular and oldest algorithms used as a basis for developing new clustering techniques [14]. The K-means algorithm categorises a number of data points into *K* clusters by calculating the summation of squared distances between every point and its nearest cluster centroids [15]. Despite its inaccuracy, the speed and simplicity of K-means are practically attractive for many scholars. Besides, K-means has several drawbacks: (i) It is difficult to predict the number of clusters. (ii) Different initial cluster centroids may result in different final clusters. (iii) K-means does not work well with datasets in different size and density. (v) Noise and outliers may misguide the clustering results. (vi) Increasing the dimension of datasets might result in bad clusters.

To overcome the shortcomings mentioned above, another clustering algorithm based on K-means, called K-means++, was proposed to generate optimal clustering results [4]. In K-means++, a smarter method is used for choosing the cluster centroids and improving the quality of clusters accordingly. Except for the initialisation process, the rest of the K-means++ algorithm is the same as the original K-means technique. That is, K-means++ is the standard K-means linked to a smarter initialisation of cluster centroids. In its initialisation procedure, K-means++ randomly chooses cluster centroids that are far from each other. This process increases the likelihood of initially collecting cluster centroids in various clusters. Because cluster centroids are obtained from the data points, each cluster centroid has certain data points at the end. In the same study, experiments conducted to compare the performance of K-means with K-means++. Though the initialisation process of K-means++ is computationally more expensive than the original K-means, K-means++ consistently outperformed K-means algorithm. However, the main drawback of K-means++ is its intrinsic sequential behaviour. That means several scans needed over the data to be able to find good initial cluster centroids. This nature limits the usability of K-means++ in big datasets.

This study [16] proposed scalable K-means++ to reduce the number of passes needed to generate proper initialisation. Lately, another technique, called expectation maximisation (EM) was introduced by [17] to maximise the overall likelihood or probability of the data, and to give a better clustering result. EM was extended from the standard K-means algorithm by computing the likelihood of the cluster membership based on one or more probability distributions. On this basis, this algorithm is suitable for estimation problems. It was also addressed that future work on EM would include its use in different and new areas, and its enhancement to improve computational structure and convergence speed.

LVQ is another crucial algorithm for clustering datasets, which was introduced by [18] to classify the patterns where each output unit is represented by a cluster. Additionally, three variants of LVQ (LVQ2, LVQ2.1 and LVQ3) were developed by [19] for building better clustering results. Nevertheless, LVQ and its variants have reference vector diverging that may decrease their recognition capability. Several studies have been conducted to solve this problem. One of the successful studies was introduced by [20] to update the reference vectors and to minimise the cost function. Experiments indicated that the recognition capability of GLVQ is high in comparison with LVQ.

Furthermore, the involvement of meta-heuristic algorithms in clustering techniques has become popular and successful in many ways. At first sight, genetic algorithm (GA) with K-means was used as a combined technique to avoid the user input requirement of $k$ and enhance the quality of the cluster by exploring good quality optimum [5]. Recently, several GA-based K-means was proposed. In these approaches, new genetic operators have been introduced to obtain faster algorithms than their predecessors and achieve better quality clustering results [21]. Meanwhile, a new clustering technique was proposed to enhance energy exploitation in wireless sensor network [22]. The proposed algorithm was combined with bat and chaotic map algorithms. The experimental results achieved with the implementation of the introduced technique had a significant impact on energy usage enhancement, increasing network lifetime, and the number of live nodes in different algorithm rounds.

One of the common GA-based algorithms is called GenClust. Two versions of GenClust were suggested by [23] and [24]. On the one hand, the first GenClust algorithm was suggested by [23] for clustering gene expression datasets. In GenClust, a new GA was used to add up two key features: (i) a smart coding of search space, which is compact, simple, and easy to update. (ii) It can naturally be employed together with data-driven validation methods. Based on experimental results on real datasets, the validity of GenClust was assured compared to other clustering techniques. On the other hand, the second version proposed by [24] as a combination of a new GA with K-means technique. Experiments conducted to evaluate the superiority of GenClust with five new algorithms on real datasets. These experiments showed a better performance of GenClust compared to its counterpart techniques. However, the second version of GenClust is not empty of shortcomings. Therefore, this algorithm was advanced by [25] namely GENCLUST++ by combining GA with K-means algorithm as a novel re-arrangement of GA operators to produce high-quality clusters. Experiments conducted and the results showed that the introduced algorithm was faster than its predecessor and could achieve a higher quality of clustering results.

After reviewing the above-mentioned clustering techniques based on meta-heuristic algorithms, we have found that these algorithms could produce better clustering results in comparison with the other classical techniques. Consequently, we propose a new cluster evolutionary algorithm in this study for clustering multiple-features and heterogeneous datasets. This suggested technique is based on the social class ranking with a combination of hybrid techniques of percentiles and quartiles, operators of the meta-heuristic algorithm, random walks in levy flight optimisation, and Euclidean distance of K-means algorithm. This newly introduced algorithm aims seven-folds: (i) Automatic selection of the number of clusters. (ii) Finding the right cluster centroids for each cluster. (iii) Preserving a balance between population diversity to avoid the algorithm to trap local and global optima. (v) Providing capability of handling scalable benchmarking problems. (vi) Providing a framework to show the sensitivity of ECA* towards different cohorts of datasets in comparison to its counterpart algorithms. (vii) Avoiding the outliers and noise to misguide the ECA* to generate good clustering results. (viii) Providing stability and robustness to our proposed algorithm.

## 3. ECA*

We integrate several new ideas to introduce a new and high-quality clustering algorithm. The first idea is to choose good quality of chromosomes instead of choosing them randomly. Using the notion of social class rankings can obtain a set of reasonably good chromosomes. Hence, ECA* uses percentile ranks for the genes of each chromosome. Based on their percentile ranks, each group of related chromosomes is allocated a cluster. Details will be presented below when we discuss the ingredients of ECA*. The second notion is the novelty of selecting and re-defining the cluster centroids ($C$) and their historical cluster centroids ($C^{old}$) using quartile and intraCluster. The value of cluster centroids ($C^{old}$) is determined using the following two steps. Initially, the cluster centroids are generated using mean quartiles, while the old cluster centroids are randomly produced between the lower and upper quartiles to avoid misguidance by noise and outliers. At the initiation of each iteration, the $C^{old}$ of each cluster is re-defined based on the value of its intraCluster. Another intervention applied to ECA* is using the operators of evolutionary algorithms. Both mutation (Mutant) and crossover are operated on the cluster centroids to pull the clusters to reach a meaningful result. At first, each cluster centroids is mutated using control parameter (F) and learning knowledge from utilising the better historical cluster centroid information (HI). HI is defined using the differences between C and $C^{old}$ if intraCluster of C is higher than intraCluster of $C^{old}$ and vice versa. F is a random walk generator using LFO. Crossover ($C^{new}$) is computed using uniform crossover between C and $C^{old}$. More significantly, these convergence operators are recombined to define new evolutionary operator called Mut-over. This new operation is used in ECA* as a recombination strategy to select between mutation and crossover based on better fitness value. Mut-over will be the value of Mutant if the interCluster of $C^{old}$ is higher than the interCluster of C and vice versa. This recombination strategy pulls the cluster centroids and their related chromosomes towards a local maximum/minimum. Besides, our algorithm allows merging clusters based on a pre-defined cluster density threshold to combine less-density clusters with others and produce the optimum number of clusters accordingly. The average intraCluster and the minimum distance between two clusters are the main criteria to decide upon their combination. Above all, cluster cost is defined as the objective function. Since ECA* is an evolutionary algorithm for performing clustering, our objective function for every problem is cluster cost. The optimal cluster cost aims to minimise the interCluster distance value and to maximise the intraCluster distance value.

To calculate these distance measures, we assume *A* and *B* are generated clusters from a clustering algorithm. *D (x, y)* is a Euclidean distance between two observations *x* and *y* belonging to *A* and B respectively. *D (x, y)* is computed using the Euclidean distance. */A/* and */B/* are the number of observations in clusters *A* and *B*, respectively. There are many definitions for calculating the interCluster distance and intraCluster distance [26]. We use these two distance measures in our study: (i) IntraCluster is the average distance between all the observations belonging to the same cluster. (ii) InterCluster is the distance between the cluster centroid and all the observations belonging to a different cluster.

IntraCluster is calculated, as shown in Expression (1) [27].

$$intraclass\ of\ A = \frac{1}{|A|.(|A|-1)} \sum_{\substack{x,y \in A \\ x \neq y}}^{n} \{d(x,y)\} \qquad (1)$$

InterCluster is computed as expressed in Equation (2) [27].

$$\text{InterCluster of (A, B)} = \frac{1}{|A|+|B|} \left\{ \sum_{x \in A} d(x, v_b) + \sum_{y \in B} d(y, v_a) \right\} \quad (2)$$

Where:

$$v_a = \frac{1}{|A|} \sum_{x \in A} x \quad (3)$$

$$v_b = \frac{1}{|B|} \sum_{y \in B} y \quad (4)$$

As it is detailed in Figure 2, ECA* includes four principal ingredients: Initialisation, Clustering-I, Mut-over, Clustering-II, and Fitness evaluation.

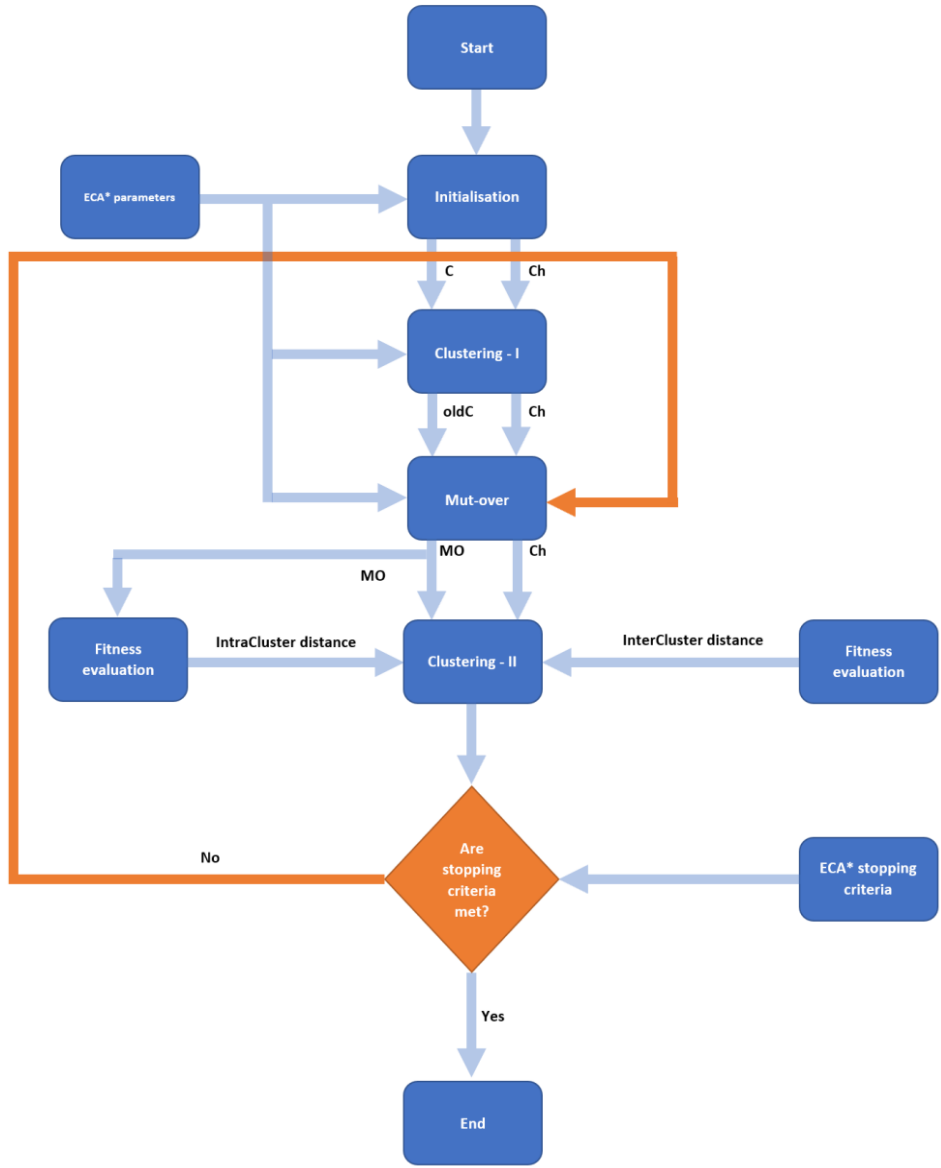

**Figure 2: A detailed flowchart of ECA***

**A. Ingredient 1: Initialisation.** The input dataset contains a number of records, and each record has several numerical or/and categorical attributes. We assume that the input data (*Dataset*) is represented as N chromosomes $Ch_i$, and each chromosome is composed of a set of genes ($G_{i0}$, $G_{i1}$, $G_{i2}$,...$G_{ij}$). For instance, $Ch_{ij} = (G_{i0}, G_{i1}, ..., G_{ij})$ has *j* number of genes.

For i=0, 1, 2,….., N  and j=0, 1, 2,….., D Where N and D are the numbers of records and attributes respectively.

At this step, the below parameters should be initialised.

1. Initialising the number of social class ranks (*S*).

2. After initialising the number of social ranks (S), the number of clusters (K) as an initial solution for the clustering problem is calculated from S power to a number of genes in each chromosome as shown in Equation (5) below:

$$K = S^j \tag{5}$$

3. Initialising the minimum cluster density threshold ($C^{dth}$).
4. Initialising the position of random walk (F).
5. Initialising the type of crossover ($C^{type}$).
6. Finding the percentile rank ($P_{ij}$) for each gene in a chromosome
7. Finding average percentile rank ($P_i$) for each chromosome.
8. Each chromosome is allocated to a cluster (K) based on its percentile rank.

**B. Ingredient 2: Clustering-I.** In this component, two main steps are calculated. (i) The number of clusters ($K^{new}$) is calculated after removing the empty clusters. (ii) The cluster centroids are generated from the actual clusters. For each group of chromosomes, $Ci$ is calculated as the cluster centroids as follows:

1) In addition to the number of new clusters ($K^{new}$), the empty clusters ($K^{empty}$), and the less-density clusters ($K^{dth}$) will be removed. The empty clusters are defined as those clusters that have no chromosomes. The actual number of clusters are computed, as presented in Equation (6).

$$K^{new} = K - K^{empty} - K^{dth} \tag{6}$$

Besides, the less-density clusters are those clusters, in which their density are less than the minimum density cluster threshold. These clusters will be merged with their nearest neighbouring clusters. The cluster density for $i$ is calculated as presented in Expression (7).

$$CD_i = \frac{Number\ of\ chromosomes\ belong\ to\ i}{N} \tag{7}$$

2) Generate the initial cluster centroids for each cluster from its chromosomes from the quartile mean of each cluster. The initial cluster centroids are determined as depicted in Equation (8).

$$C_i = mean\ quartile\ (Ch_{ij})$$

(8)

Where:

$Ch_{ij}$ is a set of chromosomes.

3) At the beginning of each iteration, the new clusters (old$C_i$) are generated randomly in between lower and upper quartiles, for each cluster of $i$, old$C_i$ is re-defined by using Equation (9) as follows:

$$oldC_i \sim U\ (lQuartile, uQuartile)$$

(9)

4) At the initiation of each iteration, intraCluster and oldIntraCluster are calculated for each cluster. Likewise, interCluster and oldInterCluster are computed for the current clustering solution.

5) After determining the historical centroids, new cluster centroids are calculated as presented in Equation (10) as follows:

$$newC_i := \begin{cases} C_i & intraCluster_i < oldIntraCluster_i \\ oldC_i & otherwise \end{cases} \tag{10}$$

**C. Ingredient 3: Mut-over.** This strategy consists of a recombination operator of mutation and crossover.

1. Mutation: Mutate each cluster centroid of $i$ in order to move to the most density part of the clusters. Mutating each cluster is calculated according to Equation (11) as shown below:

$$Mutant_i := C_i + F\ (HI) \tag{11}$$

Where:

*F*: The initial position of random walk of levy flight optimisation

*HI*: Learning knowledge from utilising the better historical information. Finding *HI* from historical information (*oldC*). *HI* is computed as presented in Equation (12) below:

$$HI := \begin{cases} oldC_i - C_i & \text{If } intraCluster_i < oldIntraCluster_i \\ C_i - oldC_i & \text{Otherwise} \end{cases} \quad (12)$$

2. Crossover: a new cluster centroid is generated from the current and old cluster centroids using a uniform crossover operator. For cluster i, the new cluster centroid (newC$_i$) is calculated as shown in Equation (13)

$$newC_i := oldC_i + C_i \quad (13)$$

3. Mut-over: it is a changeover operator between crossover and mutation of ECA* to generate the final trial of centroids between mutation and crossover on the basis of objective functions as it is described in Equation (14).

$$MO_i := \begin{cases} Mutant_i & if\ oldInterCluster_i > interCluster_i \\ newC_i & Otherwsie \end{cases} \quad (14)$$

Some cluster centroids obtained at the end of ECA*'s crossover process may overflow their search space limitations as a consequence of ECA*'s mutation strategy. The Mut-over strategy is regenerated beyond the search space limitations using Algorithm 1. ECA*'s mut-over operator is similar to BSA's boundary control mechanism [28]. Plus, ECA*'s boundary control mechanism is effectual in generating population diversity that ensures effective searches for generating good clustering and cluster centroid results.

```
Algorithm 1: Mut-over strategy of ECA*
  Input: K, MO, Mutant, C, InterClass, oldInterClass
  Output: MO
1 for i ← 0 to K do
2   if (oldInterClass_i > interClass_i) then
3     MO_i := Mutant_i;
4   end
5   else
6     MO_i := C_i;
7   end
8 end
```

**D. Ingredient 4: Clustering-II.** After the mut-over operator is generated, the clusters are merged based on their diversity, and then the cluster centroids will be re-calculated. In this step, merging between the close clusters is involved. In turn, a distance measurement between the closed clusters should be taken into account. There are a number of widely used distance measurements between clusters [29]. The most commonly used methods are minimum and maximum distance methods, centroid distance method, and cluster-average method. The minimum distance method was used in this study [30]. On these grounds, we adopt the minimum distance method in this study, in which minimum distance between clusters is considered as their merging criterion norm. Let X and Y be two clusters. Therefore, the minimum distance between them can be defined in Equation (15) [30], as shown below:

$$D_{min}(X, Y) := \min\{d(A_x, A_y)\} \quad (15)$$

Where:

$A_x \in X, A_y \in Y$

The diversity between Ci and Cj is represented in Equation (16) [30] as follows:

$$\sigma_{ij} = \min\left\{\left(D_{min}\left(C_i, C_j\right) - R(C_i)\right)\left(D_{min}\left(C_i, C_j\right) - R(C_j)\right)\right\}$$

(16)

Where:

R(ci) and R(Cj) are the average distance of the intra-cluster, $D_{min}\left(C_i, C_j\right)$ and $D_{min}\left(C_i, C_j\right)$ are the minimum distance between Ci and Cj

The merging criterion of two classes can be either: (1) If the result (□□≤□□) is less than or equal to zero, it means these two classes are close to each other, they are interconnected to a large extent. Therefore, class Ci and Cj can be merged into one class (Cij). That is after less-density class will be an empty class. (2) If □□□□, it means that the average distance of intra-class between these two classes is less than their respective shortest distance. This suggests that Ci and Cj persist as two different clusters. At last, the number of clusters will be re-calculated to remove the empty clusters as they are produced from the less-density clusters.

**F. Ingredient 5: Fitness evaluation.** The clusters are measured by fitness evaluation. The inputs are interCluster distance and intraCluster distance of the generated clusters, and the output is the fitness. If the interCluster reaches the maximum and intraCluster reaches minimum (optimal), the algorithm will stop. However, this is not realistic. The stopping criteria can be met if: (i). The algorithm reaches the initialised number of iterations. (ii). The value of interCluster and intraCluster remain un-changed during each iteration. That is, the interCluster value is not increased, and intraCluster value is not decreased during each iteration.

**Algorithm 2:** Pseudo-code of ECA*

**Input:** S, K, $C^{dth}$, F, $C^{type}$, maxcycle, P, N, D, $Ch_{ij}$
**Output:** A set of classes (MO), class centroids (C)

1 //1. INITIALISATION;
2 K = $S^D$ **for** $i \leftarrow 0$ **to** $N$ **do**
3   **for** $j \leftarrow 0$ **to** $D$ **do**
4     $Ch_{ij} = Dataset_{ij}$
5   **end**
6 **end**
7 **for** $i \leftarrow 0$ **to** $N$ **do**
8   **for** $j \leftarrow 0$ **to** $D$ **do**
9     $P_{ij}$ = Percentilisation ($Ch_{ij}$)
10   **end**
11 **end**
12 **for** $iteration \leftarrow 1$ **to** $maxcycle$ **do**
13   //2. CLASSING-I;
14   $K^{new} = K - K^{empty}$ ;
15   **for** $i \leftarrow 0$ **to** $N$ **do**
16     **if** (($Ch_i$ belongs to $P_i$) **then**
17       $C_i := MeanQuartilei(Ch_i)$;
18     **end**
19   **end**
20   // 3. MUT-OVER;
21   // 3.1. MUTATION and CROSSOVER;
22   **for** $i \leftarrow 0$ **to** $N$ **do**
23     **if** (($intraClass_i$ ¡ $oldIntraClass_i$) **then**
24       $HI_i := oldC_i - C_i$;
25     **end**
26     **else**
27       $HI_i := C_i - oldC_i$;
28     **end**
29     $Mutant = C_i + F(H_i)$;
30     $newC_i$ = uniformCrossover ($oldC_i + C_i$);
31   **end**
32   // 3.2. MUT-OVER OPERATOR: Recall Algorithm 1;
33   // 4. CLASSING-II;
34   **for** $i \leftarrow 0$ **to** $K$ **do**
35     **for** $j \leftarrow 0$ **to** $K$ **do**
36     **end**
37   **end**
38   // 5. FITNESS EVALUATION;
39   **if** ($oldInterClass_i = interClass_i$) AND ($oldIntraClass_i = intraClass_i$) **then**
40     // Export the class centroids and their observations;
41   **end**
42 **end**

The overall pseudo-code of ECA* is represented in algorithm (2). Meanwhile, The software codes in Java of ECA* can be found in [31].

## 4. Experiments

This experiment is divided into four sections. Section 4.1 presents 32 heterogenous and multiple-featured datasets used in this experiment. The internal and external evaluations of clustering results of all the techniques are discussed in Section 4.2, followed by analysing these results based on the evaluation criteria. Lastly, an operation framework is proposed to differentiate the performance sensitivity of these algorithms based on different dataset properties.

### 4.1. Benchmark datasets

In this section, we present heterogeneous, multiple-featured, and various types of datasets, which have been used to evaluate the performance of ECA* compared to its counterpart algorithms. The benchmark datasets are challenging enough for most typical clustering algorithms to be solved, but simple enough for a proper clustering algorithm to find the right cluster centroids. The benchmarking data include the following two groups of well-known datasets, which are publicly available in [2].

- Two-dimensional datasets: include A-sets, S-sets, Birch, Un-balance, and Shape sets. The basic properties of these datasets are presented in Table 2.
- N-dimensional datasets: DIM (high), and G2 sets. The basic properties of these datasets are described in Table 3.

**Table 2: Two-dimensional clustering benchmark datasets** [2]

| Dataset | Varying | type | Number of observations | Number of clusters |
|---|---|---|---|---|
| S | Overlap | S1 | 5000 | 15 |
|  |  | S2 |  |  |
|  |  | S3 |  |  |
|  |  | S4 |  |  |
| A | Number of clusters | A1 | 3000 | 20 |
|  |  | A2 | 5250 | 35 |
|  |  | A3 | 7500 | 50 |
| Birch | Structure | Birch1 | 100,000 | 100 |
| Un-balance | Both sparse and dense clusters | Un-balance | 6500 | 8 |
| Shape sets | Cluster shapes and cluster number. | Aggregation | 788 | 7 |
|  |  | Compound | 399 | 6 |
|  |  | Path-based | 300 | 3 |
|  |  | D31 | 3100 | 31 |
|  |  | R15 | 600 | 15 |
|  |  | Jain | 373 | 2 |
|  |  | Flame | 240 | 2 |

Table 3: Multi-dimensional clustering benchmark datasets [2]

| Dataset | Varying | type | Dimension- variable | Number of observations | Number of clusters |
|---|---|---|---|---|---|
| DIM (high) | Well separated clusters (cluster structure) | Dim-32 | 32 | 1024 | 16 |
| | | Dim-64 | 64 | | |
| | | Dim-128 | 128 | | |
| | | Dim-256 | 256 | | |
| | | Dim-512 | 512 | | |
| | | Dim-1024 | 1024 | | |
| G2 sets: | Cluster dimensions and overlap. | G2-16-10 | 10 | 2048 | 2 |
| | | G2-16-30 | 30 | | |
| | | G2-16-60 | 60 | | |
| | | G2-16-80 | 80 | | |
| | | G2-16-100 | 100 | | |
| | | G2-1024-10 | 10 | | |
| | | G2-1024-30 | 30 | | |
| | | G2-1024-60 | 60 | | |
| | | G2-1024-80 | 80 | | |
| | | G2-1024-100 | 100 | | |

**4.2. Experimental setup**

We conduct experiments to empirically study the performance of ECA* in comparison to its five competitive techniques (KM, KM++, EM, LVQ, and GENCLUST++). The main objectives of this experiment are three-fold: (1) Studying the performance of ECA* against its counterpart algorithms on heterogeneous and multiple-featured of datasets. (2) Evaluating the performance of ECA* against its counterpart algorithms using internal and external measuring criteria. (3) Proposing a performance framework to investigate how sensitive the performance of these algorithms on different dataset features (cluster overlap, number of clusters, cluster dimensionality, cluster structure, and cluster shape). Because clustering solutions produced by the algorithms can differ between different runs, we run ECA* with its counterpart algorithms 30 times per benchmark dataset to record their cluster quality for each run. Weka 3.9 is used to tun the five competitive techniques of ECA* per each dataset problem. We also record the average results for the (30) times clustering solutions on each dataset problem for each technique. Additionally, three pre-defined parameters are initialised for all the techniques as presented in Table 4.

Table 4: The pre-defined initial parameters

| Parameters | Initial assumption |
|---|---|
| Cluster density threshold | 0.001 |
| Alpha (random walk) | 1.001 |
| Number of social class rank | 2-10 |
| Number of iterations | 50 |

| Number of runs | 30 |
|---|---|
| Type of crossover operator | Uniform crossover |

**4.3. Evaluation of clustering results**

It has been stated in [32] that clustering evaluation and validation are considered as crucial as the clustering itself. This validation can be achieved using the internal and external validation measures.

1. External Measures: These measures require ground truth labels. Examples are adjusted rand index, centroid index, centroid similarity index, homogeneity, normalised mutual information, Fowlkes-Mallows scores, mutual information-based scores, completeness and v-measure.

2. Internal Measures: These measures do not require ground truth labels. Some of the clustering performance measures are silhouette coefficient, sum of squared error, mean squared error, Calinski-Harabasz index, and approximation ratio, and Davies Bouldin index.

As mentioned earlier, there are several internal and external measures to evaluate the clustering results. In our study, we use three internal measures and three external measures to evaluate ECA* against its counterpart techniques. For internal measures, we use the sum of squared error (SSE), mean squared error (MSE), and approximation ratio (ε- ratio). These three internal measures are efficient and sufficient for assessing the clustering results because of three-folds: (i) Using SSE to measure the squared differences between each observation with its cluster centroid and the variation with a cluster; (ii) Using MSE to measure the average of squared error, and the average squared difference between the actual value and the estimated values.; (iii) Using the *ε- ratio* to evaluate the results with the theoretical results achieved for approximation techniques. On the other hand, we use centroid index (CI), centroid similarity index (CSI), and normalised mutual information (NMI) for external measures. These three external measures are efficient and sufficient for assessing the clustering results because of three-folds: (i) Using CI to determine how many clusters are missing their centroids, or how many clusters have more than one centroid; (ii) Using CSI to measure point level differences in the matched clusters and calculate the proportional number of identical points between the matched clusters in order to provide a more precise result at the cost of interpreting the loss of value intuitively; (iii) Using NMI to give more details on the point level differences between the matched clusters.

In addition to the internal and external measures for evaluating the clustering results, we statistically evaluate ECA* with its competitive techniques (KM, KM++, EM, LVQ, GENCLUST++) in terms of their performance, which is expressed by intraCluster distance, interCluster distance, and the CPU execution time (s). For this, we have compared the ECA* with its counterpart algorithms based on their best solution (best), the worst solution (worst), and the mean solution (average) for intraCluster distance, interCluster distance, and execution time for the 30 times run on the 32 datasets.

**4.3.1. Internal measures:** The measures of internal performance evaluation rely only on the datasets themselves. These measures are all different from each other for the same objective function. We use the sum of squared error (SSE) to measure the squared differences between each observation with its cluster centroid and the variation with a cluster. If all cases with a cluster are similar, the SEE would be equal to 0. That is, the less the value of SSE represents the better work of the algorithms. For example, if one algorithm gives SSE=5.54 and another SSE=9.08, then we can presume that the former algorithm works better. The formula of SSE is shown in Equation (17) below [2]:

$$\sum_{i=1}^{N}(x_i - c_j)^2$$

(17)

Where $x_i$ is the observation data and $c_j$ is its nearest cluster centroid.

Besides, we use mean squared error (MSE) to measure the average of squared error, and the average squared difference between the actual value and the estimated values. As it is used in [2], a normalised mean squared error (nMSE) is used in this study as it is shown in Equation (18):

$$nMSE = \frac{SSE}{N.D}$$

(18)

Where:

*SSE*: the sum of squared error.

*N*: number of populations.

*D*: number of attributes in the dataset,

Nevertheless, the use of objective function values, such as SSE and nMSE could not say much about the results. Alternatively, we use the approximation ratio (*ε- ratio*) to evaluate the results with the theoretical results achieved for approximation techniques. The calculation of *ε- ratio* is presented in Equation (19) [2].

$$\varepsilon - ratio = \frac{SSE - SSE_{opt}}{SSE_{opt}} \qquad (19)$$

To calculate $SSE_{opt}$, ground centroids could be used, but their locations can vary from the optimal locations used for minimising SEE. Meanwhile, it is hard to find the $SSE_{opt}$ for A, S, Birch, Shape, G2, and dim datasets in the literature to be used in this study. The results of an algorithm on a few datasets are used in [33], but it is not sufficient to be used in this research. It is claimed that SSE should be as close as to zero [2]. Minimising SSE to zero is hard in both two and multi-dimensional datasets, and it is not practical for almost all dataset problems [34]. At the same time, it is typically not necessary to find the optimal value of SSE. For this reason, we assume that the optimal value of SEE is 0.001

**4.3.2. External measures:** The measures of external performance evaluation reply on the data that we previously have. We use three main external measures in this evaluation. As our primary measure of success, we firstly use centroid index *(CI)* to count how many clusters are missing their centroids, or how many clusters have more than one centroid. The higher value of *CI* is the less number of correct cluster centroids [35]. For example, if *CI= 0*, the result of clusters is correct, and the algorithms could solve the problem correctly. If *CI > 0*, the algorithm could not solve the problem, either some clusters are missing, or a cluster may have more than one centroid. Since *CI* measures only the cluster level differences, we also use centroid similarity index *(CSI)* to measure point level differences in the matched clusters and calculate the proportional number of identical points between the matched clusters. Accordingly, this measure provides a more precise result at the cost of interpreting the loss of value intuitively. We also use normalised mutual information *(NMI)* to give more details on the point level differences between the matched clusters. In other words, *NMI* is a normalisation of the mutual information between the cluster centroids and ground truth centroids of the same matched clusters. *NMI* scales the results between 1 (perfect correlation between the matched clusters), and 0 (no mutual information between the matched clusters). MATLAB implementation of *MNI* can be found in [36].

**4.3.3. Performance evaluation measures:** Due to their stochastic nature, evolutionary algorithms may reach worse or better solutions than solutions they have previously arrived by accident during their search for new solutions to a problem. Because of such circumstances, it is beneficial to use statistical tools to compare the problem-solving success of one algorithm with that of another. The simple statistical parameters that can be derived from the results of an algorithm solving 31 datasets K times under initial conditions. These parameters can be the best solution (best), the worst solution (worst), and the mean solution (average) to provide information about the algorithm's behaviour in solving that particular dataset problem. On that basis, we have statistically evaluated ECA* with its competitive techniques (KM, KM++, EM, LVQ, GENCLUST++) in terms of their performance. According to [37], the performance of clustering algorithms can be expressed by their intraCluster distance, interCluster distance, and the CPU execution time (s).

## 5. Results analysis

### 5.1. Performance evaluation

The results of ECA* are summarised in Table 5. It depicts the ECA* implementation over 32 datasets. The evaluation measures are based on cluster qualities *(CI, CSI, and NMI),* objective functions *(SSE, nMSE, and ε- ratio).* In most of the cases, ECA* is the sole success algorithm among its counterparts. Table 5 shows the cluster quality and objective function measures for ECA* for 30 run average.

Table 5: Cluster quality and objective function measures for ECA* for 30 run average

| Datasets | Cluster quality | | | Objective function | | |
|---|---|---|---|---|---|---|
| | CI | CSI | NMI | SSE | nMSE | ε- ratio |
| **S1** | 0 | 0.994 | 1.000 | 9.093E+12 | 9.093E+08 | 9.093E+15 |
| **S2** | 0 | 0.977 | 1.000 | 1.420E+13 | 1.420E+09 | 1.420E+16 |
| **S3** | 0 | 0.9809 | 1.000 | 1.254E+13 | 1.254E+09 | 1.254E+16 |
| **S4** | 0 | 0.872 | 1.000 | 9.103E+12 | 9.103E+08 | 9.103E+15 |
| **A1** | 0 | 1 | 1.000 | 1.215E+10 | 2.026E+06 | 1.215E+13 |
| **A2** | 0 | 0.9785 | 1.000 | 7.103E+09 | 1.184E+06 | 7.103E+12 |
| **A3** | 0 | 0.9995 | 1.000 | 2.949E+10 | 4.915E+06 | 2.949E+13 |
| **Birch1** | 0 | 0.999 | 0.989 | 7.007E+09 | 3.504E+04 | 7.007E+12 |
| **Un-balance** | 0 | 1 | 1.000 | 2.145E+11 | 1.650E+07 | 2.145E+14 |
| **Aggregation** | 0 | 1 | 0.967 | 1.024E+04 | 6.499E+00 | 1.024E+07 |
| **Compound** | 0 | 1 | 1.000 | 4.197E+03 | 5.259E+00 | 4.197E+06 |
| **Path-based** | 0 | 1 | 0.923 | 4.615E+03 | 7.691E+00 | 4.615E+06 |
| **D31** | 0 | 1 | 0.838 | 3.495E+03 | 5.637E-01 | 3.495E+06 |
| **R15** | 0 | 1 | 0.653 | 1.092E+02 | 9.099E-02 | 1.092E+05 |
| **Jain** | 0 | 1 | 1.000 | 1.493E+04 | 2.001E+01 | 1.493E+07 |
| **Flame** | 0 | 1 | 0.965 | 3.302E+03 | 6.880E+00 | 3.302E+06 |
| **Dim-32** | 0 | 0.999 | 0.807 | 1.618E+05 | 4.938E+00 | 1.618E+08 |
| **Dim-64** | 0 | 0.915 | 0.708 | 1.814E+06 | 4.324E-01 | 2.834E+07 |
| **Dim-128** | 0 | 0.907 | 0.613 | 1.958E+04 | 1.494E-01 | 1.958E+07 |
| **Dim-256** | 0 | 0.974 | 0.475 | 1.255E+04 | 4.789E-02 | 1.255E+07 |
| **Dim-512** | 0 | 0.871 | 0.330 | 2.969E+05 | 5.663E-01 | 2.969E+08 |
| **Dim-1024** | 0 | 0.934 | 0.283 | 1.992E+05 | 1.900E-01 | 1.992E+08 |
| **G2-16-10** | 0 | 1 | 0.715 | 2.048E+05 | 6.250E+00 | 3.259E+11 |
| **G2-16-30** | 0 | 1 | 0.613 | 1.825E+06 | 5.569E+01 | 1.825E+09 |
| **G2-16-60** | 0 | 0.998 | 0.571 | 1.045E+07 | 3.190E+02 | 1.045E+10 |
| **G2-16-80** | 0 | 0.997 | 0.505 | 1.701E+07 | 5.192E+02 | 1.701E+10 |
| **G2-16-100** | 0 | 0.999 | 0.566 | 3.259E+08 | 9.947E+03 | 3.879E+11 |
| **G2-1024-10** | 0 | 1 | 0.702 | 2.097E+07 | 1.000E+01 | 2.097E+10 |
| **G2-1024-30** | 0 | 0.999 | 0.584 | 2.086E+08 | 9.947E+01 | 2.086E+11 |
| **G2-1024-60** | 0 | 0.998 | 0.527 | 8.074E+08 | 3.850E+02 | 8.074E+11 |
| **G2-1024-80** | 0 | 0.997 | 0.506 | 3.040E+09 | 1.450E+03 | 3.040E+12 |
| **G2-1024-100** | 0 | 0.996 | 0.498 | 2.221E+09 | 1.059E+04 | 2.221E+13 |

Both Figures 3 and 4 elucidate the results of Table 5. Figure 3 illustrates the cluster quality measured by external measures *(CI, CSI, and NMI)* for ECA* compared to its counterpart algorithms for 30 run average. Differently, Figure 4 depicts the internal measures *(SSE, nMSE, and ε-ratio)* for ECA* compared to its counterpart algorithms for 30 run average.

Extensively, Figure 3 depicts the average values, not the scores, of the external evaluation criteria *(CI, CSI, and NMI)* for 32 numerical benchmark datasets. We also evaluate the clustering quality of ECA*, KM, KM++, EM, LVQ, and GENCLUST++ to assess the contribution of ECA*. For *CI* measure. All the algorithms, except GENCLUST++, are successful in determining the right number of clusters for the used 32 datasets. In the meantime, GENCLUST++ is successful in most of the cases. Exceptionally, it was not successful for finding the right number of centroids indexes in A3, Compound, Dim-32, Dim-128, Dim-256, Dim-512, Dim-1024, and G2-1024-80. GENCLUST++ is more successful than in two-dimensional clustering benchmark datasets in comparison to multi-dimensional clustering benchmark datasets. For CSI measure, ECA* outperforms KM, EM, KM++, LQV and GENGLUST++. In a few cases, the algorithms measured 1 or approximate to 1 for *CSI* value.

Overall, ECA* is successful in having the optimal value of *CSI* for all the datasets, except S4. Alongside with its success, ECA*'s counterpart algorithms are successful in some datasets. Mainly, KM is successful in almost of shape datasets, including Aggregation, Compound, Path-based, D31, R15, Jain, and Flame, and three of the multi-dimensional clustering benchmark datasets, including G2-16-10, G2-16-30, and G2-1024-100. After that, EM is the third successful technique, in which a winner in six datasets (S2, S4, Compound, R15, Jain, and G2-1024-100), followed by KM++ which is the fourth winner in four datasets (R15, Dim-256, Dim-256, and G2-1024-100). Meanwhile, LVQ and GENCLUST++ are the least successful winner in only two datasets. Lastly, for the cluster quality measured by NMI for ECA* and its competitive algorithms for all the used datasets, ECA* cluster centroid results have the best correlation between its cluster centroids and ground truth centroids of the same matched clusters. Afterwards, KM has relatively good relations of its cluster centroids with the ground truth centroids of the same matched clusters, particularly, for the two-dimensional benchmark datasets. In terms of this correlation, KM is followed by KM++, EM, and LVQ. It needs to mention that GENCLUST++ records the least correlation for almost all the current datasets.

Generally, the results show that ECA* mostly overcomes its existing algorithms for cluster quality evaluation criteria. Correctly, ECA*, KM, KM++, EM, and LVQ perform well for determining the correct number of centroid indexes for all the 32 datasets, but GENCLUS++ does not find the right number of centroid indexes in all cases. Regarding *CSI*, ECA* outperforms its counterpart techniques. After that, EM performs better than the other four algorithms, followed by LVQ, KM++, KM, and GENCLUST++, respectively. For evaluating these techniques based on normalised mutual information (NMI), both ECA* and KM perform pretty much the same. However, ECA* slightly outperforms KM for finding a better quality of clusters. After these two techniques, KM++, EM, and LVQ perform quite similar in providing better cluster quality. Finally, GENCLUST++ is the least performing algorithm among its counterparts.

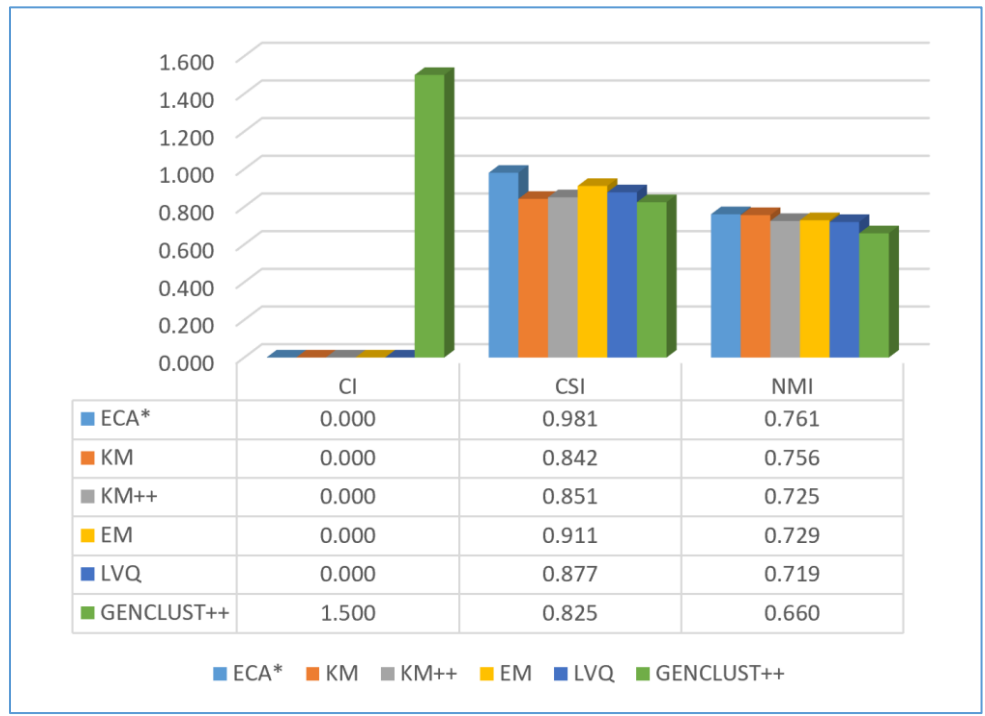

**Figure 3: The average values of the internal evaluation criteria (CI, CSI, and NMI) for six algorithms in 31 numerical datasets**

Furthermore, Figure 4 presents the average values of the internal evaluation criteria (SSE, nMSE, and ε- ratio) for 32 numerical datasets. SSE is used to measure the squared differences between each data observation with its cluster centroid. In this comparison, ECA* performs well in most of the datasets, and it is the winner in 25 datasets. On the other hand, the counterpart algorithms of ECA* are the winner in a few numbers of datasets. Therefore, ECA* performs well compared to its competitive algorithms in the current datasets. In addition to SSE, the comparison of internal cluster evaluation measured by nMSE for ECA* and its counterpart algorithms, ECA* performs well compared to its counterpart techniques in most of the datasets. After that, GENCLUS++ is the second winner that has good results in four datasets (S2, Aggregation, Flame, and G2-16-80). On the other hand, KM is not a winner in all cases, whereas KM++ is the winner in two datasets (A3, and G2-1024-100). It is necessary to mention that KM++, EM, and LVQ are the winners for G2-1024-100 dataset. Also, EM is the winner for S1, and LVQ is the winner for two other datasets (Dim-512, and G2-16-100).

Finally, the comparison of internal cluster evaluation measured by ε- ratio for ECA* and its counterpart algorithms, ECA* performs well compared to its counterpart techniques in most of the datasets. After that, EM is the second winner that has good results in three datasets (S3, R15, and G2-16-100). Meanwhile, KM++, LVQ, and GENCLUS++ are the winner in only one case, whereas KM does not win in any of the 32 datasets. Overall, the results indicate that ECA* overcomes its competitive algorithms in all the mentioned internal evaluation measures. ECA* records the minimum value of SSE followed by EM, KM++, KM, GENCLUS++, and LVQ. For nMSE, ECA* outperforms other techniques, while EM marginally performs the same as ECA*. Succeeded by these two algorithms, these techniques GENCLUS++, KM++, and KM perform relatively similar in nMSE. Lastly, LVQ has the least performed algorithm on the list. Once more, ECA* records the minimum value for ε- ratio, followed by EM. That is, ECA* and EM have the best objective function measures among their counterpart techniques once their cluster results are compared with the theoretical results achieved for approximation techniques. KM++, GENCLUS++, and KM are in the second position for objective function measured by ε- ratio. Incidentally, LVQ records the worst objective function measured by ε- ratio among its counterpart algorithms.

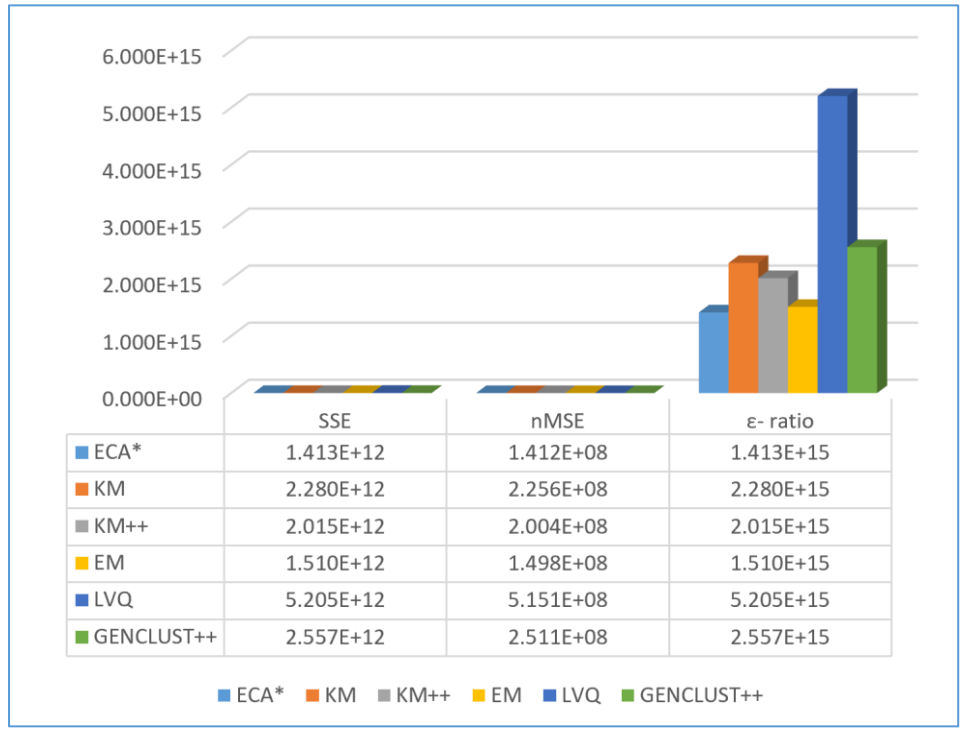

**Figure 4: The average values of the external evaluation criteria (SSE, nMSE, and ε- ratio) for six algorithms in 31 numerical datasets**

## 5.2. Statistical performance analysis

For this study, we have compared the ECA* with its counterpart algorithms based on their best solution (best), the worst solution (worst), and the mean solution (average) for intraCluster distance, interCluster distance, and execution time for the 30 times run on the 32 datasets. The results of statistical performance analysis of ECA* in comparison with its counterpart algorithms are presented in Tables 6, 7, 8.

Table 6 represents the basic statistics of the 30-solutions obtained by ECA*, KM, KM++, EM, LVQ, GENCLUST++ for intraCluster distance (Best: the best-solution, Worst: the worst solution, average: mean-solution).

**Table 6: Basic statistics of the 30-solutions obtained by ECA*, KM, KM++, EM, LVQ, GENCLUST++ for intraCluster distance (Best: the best-solution, Worst: the worst solution, average: mean-solution)**

| Datasets | Statistics | ECA* | KM | KM++ | EM | LVQ | GENCLUST++ | Winner |
|---|---|---|---|---|---|---|---|---|
| S1 | Best | 2.102E+07 | 1.064E+06 | 1.128E+06 | 1.279E+06 | 1.364E+06 | 1.364E+06 | ECA* |
|  | Worst | 1.550E+06 | 1.347E+04 | 1.511E+04 | 1.148E+04 | 8.744E+05 | 1.312E+05 | ECA* |
|  | Average | 7.119E+06 | 5.459E+05 | 4.817E+05 | 4.419E+05 | 1.000E+06 | 5.812E+05 | ECA* |
| S2 | Best | 3.584E+06 | 3.052E+07 | 1.769E+06 | 1.517E+06 | 2.871E+06 | 3.680E+06 | GENCLUST++ |
|  | Worst | 1.615E+06 | 3.375E+05 | 1.028E+06 | 7.524E+05 | 1.563E+06 | 1.143E+06 | ECA* |
|  | Average | 2.891E+06 | 1.181E+06 | 1.329E+06 | 1.299E+06 | 2.623E+06 | 1.644E+06 | ECA* |
| S3 | Best | 2.004E+07 | 2.451E+06 | 1.746E+06 | 1.274E+06 | 2.652E+06 | 1.677E+06 | ECA* |
|  | Worst | 1.634E+06 | 9.794E+05 | 1.203E+05 | 1.706E+05 | 1.818E+05 | 2.400E+05 | ECA* |
|  | Average | 7.692E+06 | 2.082E+05 | 2.841E+05 | 2.672E+05 | 2.746E+05 | 1.079E+05 | ECA* |
| S4 | Best | 7.082E+06 | 7.482E+05 | 1.843E+06 | 6.320E+06 | 4.697E+06 | 3.183E+06 | ECA* |
|  | Worst | 1.644E+06 | 5.939E+05 | 5.184E+05 | 2.145E+05 | 2.285E+05 | 1.618E+05 | ECA* |
|  | Average | 3.701E+06 | 2.383E+05 | 1.816E+05 | 3.665E+06 | 2.451E+06 | 2.979E+06 | ECA* |
| A1 | Best | 7.728E+04 | 5.958E+04 | 6.446E+04 | 2.978E+04 | 7.884E+04 | 5.506E+04 | LVQ |
|  | Worst | 9.812E+03 | 8.496E+03 | 9.389E+03 | 3.545E+03 | 1.134E+04 | 6.649E+03 | LVQ |
| | Average | 4.283E+04 | 1.616E+04 | 2.253E+04 | 4.108E+04 | 4.382E+04 | 3.915E+04 | LVQ |
| A2 | Best | 7.195E+05 | 6.069E+05 | 5.487E+05 | 2.722E+05 | 5.554E+05 | 5.574E+05 | ECA* |
| | Worst | 1.679E+05 | 1.315E+05 | 1.441E+05 | 1.420E+05 | 1.501E+05 | 1.789E+05 | GENCLUST++ |
| | Average | 4.253E+05 | 3.512E+04 | 3.284E+04 | 3.348E+04 | 3.712E+04 | 4.003E+04 | ECA* |
| A3 | Best | 2.118E+06 | 2.080E+05 | 2.015E+05 | 2.027E+05 | 2.063E+05 | 2.021E+05 | ECA* |
| | Worst | 1.824E+05 | 1.812E+03 | 1.842E+03 | 1.933E+04 | 1.959E+04 | 2.001E+04 | ECA* |
| | Average | 5.195E+05 | 5.115E+04 | 6.939E+04 | 7.123E+04 | 8.924E+04 | 2.951E+04 | ECA* |
| Birch1 | Best | 9.572E+06 | 7.832E+05 | 7.134E+05 | 8.126E+05 | 8.235E+05 | 8.724E+05 | ECA* |
| | Worst | 4.208E+06 | 3.414E+05 | 3.564E+05 | 4.258E+05 | 8.116E+05 | 5.692E+05 | ECA* |
| | Average | 7.184E+06 | 4.844E+05 | 4.632E+05 | 7.786E+05 | 8.187E+05 | 6.752E+05 | ECA* |
| Un-balance | Best | 2.410E+06 | 3.182E+05 | 4.112E+05 | 8.067E+05 | 3.256E+05 | 7.713E+05 | ECA* |
| | Worst | 1.592E+05 | 1.329E+05 | 1.445E+05 | 1.636E+05 | 1.135E+05 | 1.404E+05 | EM |
| | Average | 6.247E+05 | 2.585E+05 | 2.613E+05 | 4.517E+05 | 2.618E+05 | 4.252E+05 | ECA* |
| Aggregation | Best | 5.661E+01 | 4.277E+01 | 4.333E+01 | 4.024E+01 | 3.383E+01 | 3.383E+01 | ECA* |
| | Worst | 3.199E+01 | 1.666E+01 | 1.049E+01 | 1.475E+00 | 5.156E+00 | 8.263E+00 | ECA* |
| | Average | 4.419E+01 | 3.313E+01 | 3.413E+01 | 3.219E+01 | 4.313E+01 | 4.222E+01 | ECA* |
| Compound | Best | 5.513E+01 | 2.209E+01 | 1.884E+01 | 5.212E+01 | 1.470E+01 | 3.947E+01 | EM |
| | Worst | 3.554E+01 | 8.646E+00 | 5.190E+00 | 1.671E+00 | 3.968E+00 | 3.619E+01 | GENCLUST++ |
| | Average | 4.701E+01 | 3.913E+01 | 3.982E+01 | 4.121E+01 | 3.813E+01 | 3.713E+01 | ECA* |
| Path-based | Best | 4.818E+01 | 2.113E+01 | 2.183E+01 | 2.792E+01 | 3.245E+01 | 2.623E+01 | ECA* |
| | Worst | 2.842E+01 | 1.239E+01 | 1.027E+01 | 1.415E+01 | 1.623E+01 | 1.413E+01 | ECA* |
| | Average | 3.813E+01 | 1.612E+01 | 1.572E+01 | 2.225E+01 | 2.813E+01 | 2.513E+01 | ECA* |
| D31 | Best | 6.124E+01 | 4.867E+01 | 3.216E+01 | 3.151E+01 | 4.934E+01 | 3.614E+01 | ECA* |
| | Worst | 3.782E+01 | 2.462E+01 | 1.751E+01 | 2.144E+01 | 3.962E+01 | 2.644E+01 | LVQ |
| | Average | 5.014E+01 | 3.224E+01 | 2.712E+01 | 2.813E+01 | 4.313E+01 | 2.913E+01 | ECA* |
| R15 | Best | 1.274E+01 | 9.993E+00 | 8.904E+00 | 7.327E+00 | 9.450E+00 | 9.655E+00 | ECA* |
| | Worst | 8.850E+00 | 6.561E+00 | 5.679E+00 | 2.332E+00 | 5.459E+00 | 4.145E+00 | ECA* |
| | Average | 9.419E+00 | 7.129E+00 | 7.918E+00 | 6.813E+00 | 7.272E+00 | 7.166E+00 | ECA* |
| Jain | Best | 5.110E+01 | 3.417E+01 | 2.159E+01 | 3.639E+01 | 3.644E+01 | 4.129E+01 | ECA* |
| | Worst | 2.505E+01 | 1.751E+01 | 1.524E+01 | 1.433E+01 | 1.580E+01 | 1.663E+01 | ECA* |
| | Average | 4.018E+01 | 2.612E+01 | 1.913E+01 | 2.913E+01 | 2.813E+01 | 3.119E+01 | ECA* |
| Flame | Best | 2.602E+01 | 1.395E+01 | 9.065E+00 | 2.480E+01 | 1.515E+01 | 1.968E+01 | ECA* |
| | Worst | 1.276E+01 | 5.557E+00 | 4.507E+00 | 1.376E+01 | 8.835E+00 | 2.755E+00 | EM |
| | Average | 1.918E+01 | 7.013E+00 | 7.128E+00 | 1.612E+01 | 1.213E+01 | 1.113E+01 | ECA* |
| Dim-32 | Best | 2.198E+04 | 1.892E+04 | 2.162E+04 | 1.802E+04 | 1.735E+04 | 1.635E+04 | ECA* |
| | Worst | 9.777E+03 | 2.793E+03 | 9.576E+03 | 3.436E+03 | 5.127E+03 | 7.302E+03 | ECA* |
| | Average | 1.292E+04 | 1.625E+04 | 9.730E+03 | 1.050E+04 | 7.386E+03 | 1.534E+04 | ECA* |
| Dim-64 | Best | 9.183E+03 | 2.577E+03 | 5.813E+03 | 8.718E+03 | 3.130E+03 | 8.277E+02 | ECA* |
| | Worst | 3.580E+03 | 1.862E+02 | 2.570E+02 | 2.550E+02 | 1.050E+02 | 2.695E+02 | ECA* |
| | Average | 6.298E+03 | 1.772E+03 | 2.956E+03 | 5.319E+03 | 2.244E+03 | 6.770E+02 | ECA* |
| Dim-128 | Best | 9.842E+03 | 9.456E+03 | 8.828E+03 | 7.129E+03 | 7.624E+03 | 5.897E+03 | ECA* |
| | Worst | 5.506E+03 | 4.291E+03 | 3.087E+03 | 1.777E+03 | 4.499E+03 | 2.332E+03 | ECA* |
| | Average | 7.128E+03 | 6.397E+03 | 3.255E+03 | 1.964E+03 | 7.065E+03 | 2.563E+03 | ECA* |
| Dim-256 | Best | 1.384E+04 | 9.076E+03 | 1.278E+04 | 1.208E+04 | 1.434E+04 | 8.702E+03 | LVQ |
| | Worst | 8.582E+03 | 2.646E+03 | 1.563E+03 | 8.052E+03 | 8.509E+03 | 8.559E+03 | ECA* |
| | Average | 1.012E+04 | 7.066E+03 | 9.115E+03 | 9.128E+03 | 9.090E+03 | 8.613E+03 | ECA* |

| | | | | | | | | Winner |
|---|---|---|---|---|---|---|---|---|
| Dim-512 | Best | 7.813E+05 | 9.986E+04 | 7.718E+05 | 6.596E+05 | 5.544E+04 | 3.474E+05 | ECA* |
| | Worst | 1.920E+04 | 1.174E+04 | 8.548E+03 | 1.102E+04 | 1.750E+04 | 3.006E+03 | ECA* |
| | Average | 3.817E+05 | 6.996E+04 | 6.013E+05 | 6.308E+05 | 2.870E+04 | 3.283E+05 | ECA* |
| Dim-1024 | Best | 1.194E+05 | 1.099E+05 | 1.087E+05 | 6.821E+04 | 8.057E+04 | 9.706E+04 | ECA* |
| | Worst | 6.298E+04 | 2.007E+03 | 2.588E+04 | 2.472E+04 | 2.571E+04 | 6.321E+04 | GENCLUST++ |
| | Average | 8.714E+04 | 4.895E+04 | 9.596E+04 | 5.649E+04 | 7.978E+04 | 7.347E+04 | ECA* |
| G2-16-10 | Best | 8.913E+04 | 8.020E+04 | 6.895E+04 | 5.752E+04 | 3.078E+04 | 6.747E+04 | ECA* |
| | Worst | 4.764E+03 | 3.202E+03 | 4.567E+03 | 3.926E+03 | 1.365E+03 | 3.576E+03 | ECA* |
| | Average | 6.125E+04 | 6.450E+04 | 1.146E+04 | 4.245E+04 | 7.522E+03 | 1.116E+04 | ECA* |
| G2-16-30 | Best | 1.119E+04 | 1.064E+04 | 1.102E+04 | 6.252E+03 | 8.644E+03 | 9.184E+03 | ECA* |
| | Worst | 6.701E+03 | 4.456E+02 | 5.924E+02 | 3.763E+03 | 2.766E+03 | 4.680E+02 | ECA* |
| | Average | 9.183E+03 | 1.944E+03 | 9.221E+03 | 5.339E+03 | 4.560E+03 | 4.865E+03 | ECA* |
| G2-16-60 | Best | 1.392E+04 | 1.024E+04 | 1.119E+04 | 1.148E+04 | 1.128E+04 | 1.251E+04 | ECA* |
| | Worst | 9.264E+03 | 8.246E+03 | 7.951E+03 | 7.749E+03 | 6.182E+03 | 5.233E+03 | ECA* |
| | Average | 6.184E+04 | 9.835E+03 | 1.059E+04 | 9.427E+03 | 1.018E+04 | 1.007E+04 | ECA* |
| G2-16-80 | Best | 8.917E+04 | 5.709E+04 | 5.975E+04 | 4.567E+04 | 5.790E+04 | 6.855E+04 | ECA* |
| | Worst | 2.909E+04 | 1.010E+04 | 2.608E+04 | 3.038E+04 | 7.941E+03 | 2.265E+04 | EM |
| | Average | 6.287E+04 | 4.864E+04 | 5.779E+04 | 4.322E+04 | 4.773E+04 | 4.062E+04 | ECA* |
| G2-16-100 | Best | 1.180E+04 | 8.027E+03 | 8.608E+03 | 8.494E+03 | 1.024E+04 | 1.297E+04 | GENCLUST++ |
| | Worst | 5.746E+03 | 3.351E+03 | 9.299E+02 | 4.024E+03 | 5.888E+03 | 2.355E+02 | LVQ |
| | Average | 8.915E+03 | 6.556E+03 | 2.114E+03 | 5.690E+03 | 8.103E+03 | 5.073E+03 | ECA* |
| G2-1024-10 | Best | 9.725E+05 | 3.873E+05 | 8.171E+05 | 5.413E+05 | 8.709E+05 | 8.293E+05 | ECA* |
| | Worst | 3.287E+05 | 1.867E+05 | 2.019E+05 | 1.944E+05 | 1.589E+05 | 8.396E+04 | ECA* |
| | Average | 6.127E+05 | 3.123E+05 | 2.151E+05 | 3.777E+05 | 4.572E+05 | 7.113E+05 | ECA* |
| G2-1024-30 | Best | 8.938E+05 | 8.154E+05 | 5.292E+05 | 8.782E+05 | 4.720E+05 | 8.622E+05 | GENCLUST++ |
| | Worst | 4.714E+05 | 3.217E+05 | 8.889E+04 | 1.203E+05 | 3.941E+05 | 2.812E+03 | ECA* |
| | Average | 5.769E+05 | 6.380E+05 | 3.535E+05 | 4.884E+05 | 4.489E+05 | 1.342E+05 | ECA* |
| G2-1024-60 | Best | 2.298E+06 | 1.846E+06 | 2.041E+06 | 1.536E+06 | 1.350E+06 | 2.091E+06 | ECA* |
| | Worst | 9.735E+05 | 3.087E+05 | 4.736E+05 | 6.935E+05 | 9.338E+05 | 8.763E+04 | LVQ |
| | Average | 6.002E+06 | 6.633E+05 | 1.302E+06 | 1.047E+06 | 1.134E+06 | 1.104E+06 | ECA* |
| G2-1024-80 | Best | 2.626E+06 | 1.908E+06 | 1.085E+06 | 2.359E+06 | 1.682E+06 | 2.161E+06 | ECA* |
| | Worst | 9.416E+05 | 7.958E+05 | 4.974E+05 | 3.855E+05 | 4.501E+05 | 9.647E+05 | GENCLUST++ |
| | Average | 6.393E+05 | 1.885E+06 | 6.348E+05 | 2.284E+06 | 8.664E+05 | 1.398E+06 | ECA* |
| G2-1024-100 | Best | 1.133E+06 | 1.050E+06 | 8.348E+05 | 6.196E+05 | 9.160E+05 | 8.352E+05 | ECA* |
| | Worst | 3.671E+05 | 3.426E+05 | 3.647E+05 | 1.718E+05 | 3.846E+05 | 3.519E+05 | LVQ |
| | Average | 6.451E+05 | 4.143E+05 | 6.826E+05 | 1.980E+05 | 8.406E+05 | 7.383E+05 | ECA* |

Table7 represents the basic statistics of the 30-solutions obtained by ECA*, KM, KM++, EM, LVQ, GENCLUST++ for interCluster distance (Best: the best-solution, Worst: the worst solution, average: mean-solution).

**Table 7: Basic statistics of the 30-solutions obtained by ECA*, KM, KM++, EM, LVQ, GENCLUST++ for interCluster distance (Best: the best-solution, Worst: the worst solution, average: mean-solution)**

| Datasets | Statistics | ECA* | KM | KM++ | EM | LVQ | GENCLUST++ | Winner |
|---|---|---|---|---|---|---|---|---|
| S1 | Best | 7.010E+05 | 8.414E+06 | 7.820E+06 | 9.000E+05 | 9.482E+05 | 8.170E+05 | ECA* |
| | Worst | 1.034E+06 | 1.715E+07 | 1.373E+07 | 1.998E+06 | 1.847E+06 | 1.643E+06 | ECA* |
| | Average | 9.873E+05 | 12876803 | 12147609 | 1096204 | 1580764 | 1082023 | ECA* |
| S2 | Best | 5.582E+04 | 5.676E+05 | 2.103E+05 | 7.123E+05 | 6.553E+05 | 2.630E+05 | ECA* |
| | Worst | 9.467E+05 | 1.964E+06 | 2.802E+06 | 3.236E+06 | 1.169E+07 | 6.576E+06 | ECA* |
| | Average | 7.398E+04 | 1465879 | 1397178 | 1190439 | 9984946 | 3307972 | ECA* |
| S3 | Best | 1.859E+05 | 2.680E+06 | 9.405E+05 | 1.849E+06 | 5.970E+06 | 4.603E+06 | ECA* |
| | Worst | 6.945E+05 | 3.323E+07 | 4.549E+07 | 6.232E+07 | 4.151E+07 | 8.093E+07 | ECA* |
| | Average | 3.398E+05 | 28869705 | 38299711 | 45866919 | 38591421 | 70717731 | ECA* |
| S4 | Best | 6.865E+04 | 8.041E+05 | 9.602E+05 | 9.540E+05 | 6.612E+04 | 8.879E+05 | LVQ |
| | Worst | 1.034E+06 | 1.032E+07 | 1.210E+07 | 1.008E+07 | 1.804E+06 | 8.863E+06 | ECA* |
| | Average | 1.214E+05 | 5.807E+06 | 8.234E+06 | 7.816E+06 | 2.062E+05 | 2.658E+06 | ECA* |
| A1 | Best | 2.525E+03 | 8.929E+04 | 9.608E+04 | 2.948E+04 | 5.790E+04 | 1.050E+05 | ECA* |
| | Worst | 1.518E+04 | 8.882E+05 | 1.937E+06 | 1.788E+06 | 1.661E+06 | 8.180E+05 | ECA* |
| | Average | 7.871E+03 | 7.401E+05 | 4.765E+05 | 1.127E+06 | 1.521E+06 | 7.867E+05 | ECA* |
| A2 | Best | 5.433E+03 | 1.525E+05 | 3.616E+04 | 1.922E+05 | 1.393E+05 | 2.257E+04 | ECA* |
| | Worst | 2.176E+04 | 2.870E+05 | 3.106E+05 | 3.276E+05 | 3.376E+05 | 3.119E+05 | ECA* |
| | Average | 9.452E+03 | 2.837E+05 | 2.503E+05 | 2.802E+05 | 2.888E+05 | 1.825E+05 | ECA* |
| A3 | Best | 4.817E+03 | 2.669E+04 | 7.156E+04 | 5.160E+04 | 7.449E+04 | 4.794E+03 | GENCLUST++ |
| | Worst | 2.074E+04 | 3.199E+05 | 2.330E+05 | 1.059E+05 | 8.345E+04 | 5.005E+04 | ECA* |
| | Average | 8.512E+03 | 1.636E+05 | 8.714E+04 | 7.775E+04 | 1.489E+04 | 4.921E+04 | ECA* |
| Birch1 | Best | 2.748E+02 | 6.043E+02 | 5.488E+02 | 3.559E+02 | 3.238E+02 | 3.487E+02 | ECA* |
| | Worst | 3.983E+02 | 7.017E+02 | 3.439E+02 | 4.750E+02 | 3.886E+02 | 4.217E+02 | LVQ |
| | Average | 3.218E+02 | 6.666E+02 | 4.126E+02 | 4.181E+02 | 3.304E+02 | 3.918E+02 | ECA* |
| Un-balance | Best | 5.816E+03 | 6.556E+04 | 3.794E+04 | 6.457E+03 | 6.394E+04 | 9.811E+03 | ECA* |
| | Worst | 7.747E+04 | 7.914E+05 | 6.484E+05 | 8.811E+04 | 9.661E+04 | 3.225E+05 | ECA* |
| | Average | 9.982E+03 | 6.914E+04 | 7.837E+04 | 2.251E+04 | 7.398E+04 | 2.487E+04 | ECA* |
| Aggregation | Best | 9.283E+00 | 1.604E+01 | 1.058E+01 | 9.022E+00 | 1.578E+01 | 1.085E+01 | EM |
| | Worst | 1.567E+01 | 2.251E+01 | 4.217E+01 | 2.747E+01 | 4.214E+01 | 3.184E+01 | ECA* |
| | Average | 1.098E+00 | 1.948E+01 | 2.703E+01 | 1.324E+01 | 2.939E+01 | 1.501E+01 | ECA* |
| Compound | Best | 4.923E+00 | 1.044E+01 | 9.812E+00 | 9.386E+00 | 1.058E+01 | 7.462E+00 | ECA* |
| | Worst | 8.796E+00 | 1.574E+01 | 1.844E+01 | 1.548E+01 | 1.682E+01 | 1.448E+01 | ECA* |
| | Average | 5.947E+00 | 1.220E+01 | 1.399E+01 | 1.213E+01 | 1.189E+01 | 1.000E+01 | ECA* |
| Path-based | Best | 1.002E+01 | 1.622E+01 | 1.612E+01 | 1.465E+01 | 9.388E+00 | 1.732E+01 | LVQ |
| | Worst | 1.852E+01 | 2.924E+01 | 2.889E+01 | 2.738E+01 | 2.233E+01 | 2.817E+01 | ECA* |
| | Average | 1.349E+01 | 2.593E+01 | 2.404E+01 | 2.124E+01 | 1.323E+01 | 2.321E+01 | LVQ |
| D31 | Best | 2.339E+00 | 3.937E+00 | 4.085E+00 | 4.792E+00 | 3.878E+00 | 4.659E+00 | ECA* |
| | Worst | 4.513E+00 | 7.458E+00 | 7.293E+00 | 6.680E+00 | 5.832E+00 | 5.441E+00 | ECA* |
| | Average | 3.149E+00 | 5.014E+00 | 4.983E+00 | 5.493E+00 | 4.492E+00 | 4.912E+00 | ECA* |
| R15 | Best | 8.320E-01 | 1.776E+00 | 1.630E+00 | 9.706E-01 | 1.172E+00 | 7.929E-01 | GENCLUST++ |
| | Worst | 1.455E+00 | 2.428E+00 | 3.654E+00 | 2.550E+00 | 2.804E+00 | 2.520E+00 | ECA* |
| | Average | 1.119E+00 | 2.159E+00 | 2.019E+00 | 1.792E+00 | 1.794E+00 | 1.398E+00 | ECA* |
| Jain | Best | 1.902E+01 | 2.701E+01 | 2.463E+01 | 2.143E+01 | 2.463E+01 | 2.064E+01 | ECA* |
| | Worst | 2.883E+01 | 4.708E+01 | 4.430E+01 | 4.100E+01 | 3.902E+01 | 4.869E+01 | ECA* |
| | Average | 2.333E+00 | 3.501E+01 | 3.401E+01 | 2.973E+01 | 3.101E+01 | 3.210E+01 | ECA* |
| Flame | Best | 8.235E-01 | 6.227E+00 | 4.683E+00 | 2.668E+00 | 2.890E+00 | 3.128E+00 | ECA* |
| | Worst | 1.278E+01 | 1.663E+01 | 1.600E+01 | 1.535E+01 | 1.469E+01 | 1.761E+01 | ECA* |
| | Average | 1.005E+00 | 9.008E+00 | 8.019E+00 | 7.193E+00 | 8.188E+00 | 9.019E+00 | ECA* |
| Dim-32 | Best | 1.895E+00 | 2.9941055 | 3.8239462 | 2.103E+00 | 2.321E+00 | 2.9377751 | ECA* |
| | Worst | 2.993E+00 | 5.8629892 | 5.7921008 | 2.898E+00 | 4.3506978 | 4.3808699 | EM |

23| | | | | | | | | |
|---|---|---|---|---|---|---|---|---|
| | Average | 2.206E+00 | 3.716E+00 | 4.402E+00 | 2.400E+00 | 3.392E+00 | 3.409E+00 | ECA* |
| Dim-64 | Best | 1.923E-01 | 2.601E-01 | 2.559E-01 | 2.172E-01 | 2.096E-01 | 2.192E-01 | ECA* |
| | Worst | 2.635E-01 | 3.882E-01 | 3.626E-01 | 3.348E-01 | 3.233E-01 | 3.398E-01 | ECA* |
| | Average | 2.232E-01 | 3.149E-01 | 3.002E-01 | 2.812E-01 | 2.509E-01 | 2.640E-01 | ECA* |
| Dim-128 | Best | 2.294E-01 | 2.676E-01 | 3.244E-01 | 2.131E-01 | 2.776E-01 | 2.562E-01 | EM |
| | Worst | 3.211E-01 | 4.410E-01 | 4.227E-01 | 3.952E-01 | 3.421E-01 | 3.560E-01 | ECA* |
| | Average | 2.751E-01 | 3.402E-01 | 2.912E-01 | 3.008E-01 | 2.912E-01 | 3.094E-01 | ECA* |
| Dim-256 | Best | 1.924E-01 | 3.068E-01 | 2.928E-01 | 2.356E-01 | 1.869E-01 | 2.177E-01 | LVQ |
| | Worst | 3.375E-01 | 5.579E-01 | 5.350E-01 | 3.816E-01 | 3.814E-01 | 3.802E-01 | ECA* |
| | Average | 2.399E-01 | 4.295E-01 | 3.901E-01 | 3.018E-01 | 2.602E-01 | 2.902E-01 | ECA* |
| Dim-512 | Best | 2.578E-03 | 2.886E-02 | 2.485E-02 | 2.830E-02 | 2.549E-03 | 2.276E-03 | LVQ |
| | Worst | 1.507E-02 | 4.512E-02 | 3.176E-02 | 3.653E-02 | 4.759E-02 | 3.611E-02 | ECA* |
| | Average | 5.234E-03 | 3.186E-02 | 2.492E-02 | 2.391E-02 | 1.494E-02 | 2.185E-02 | ECA* |
| Dim-1024 | Best | 3.918E-03 | 8.983E-03 | 6.141E-02 | 7.717E-02 | 8.605E-02 | 8.922E-02 | ECA* |
| | Worst | 9.407E-03 | 7.747E-02 | 1.457E-02 | 2.682E-02 | 2.358E-03 | 5.275E-03 | ECA* |
| | Average | 4.793E-03 | 6.129E-02 | 4.591E-02 | 5.013E-02 | 1.583E-02 | 1.937E-03 | ECA* |
| G2-16-10 | Best | 2.424E+01 | 4.216E+01 | 4.845E+01 | 4.649E+01 | 3.075E+01 | 6.000E+01 | ECA* |
| | Worst | 5.345E+01 | 7.251E+01 | 7.555E+01 | 6.835E+01 | 6.757E+01 | 7.820E+01 | ECA* |
| | Average | 3.710E+01 | 5.104E+01 | 4.908E+01 | 5.800E+01 | 4.509E+01 | 6.092E+01 | ECA* |
| G2-16-30 | Best | 6.402E+01 | 8.753E+01 | 8.135E+01 | 8.159E+01 | 9.475E+01 | 8.910E+01 | ECA* |
| | Worst | 1.188E+02 | 1.756E+02 | 2.080E+02 | 1.759E+02 | 1.907E+02 | 1.096E+02 | GENCLUST++ |
| | Average | 7.988E+01 | 1.108E+02 | 9.738E+01 | 9.402E+01 | 1.028E+02 | 1.009E+02 | ECA* |
| G2-16-60 | Best | 7.423E+01 | 1.022E+02 | 1.426E+02 | 9.435E+01 | 9.384E+01 | 9.766E+01 | ECA* |
| | Worst | 1.838E+02 | 3.434E+02 | 2.422E+02 | 2.572E+02 | 1.880E+02 | 1.433E+02 | GENCLUST++ |
| | Average | 1.003E+02 | 2.780E+02 | 2.029E+02 | 1.898E+02 | 1.420E+02 | 1.280E+02 | ECA* |
| G2-16-80 | Best | 1.699E+02 | 2.050E+02 | 2.068E+02 | 2.070E+02 | 2.029E+02 | 2.048E+02 | ECA* |
| | Worst | 3.956E+02 | 3.020E+02 | 2.763E+02 | 3.177E+02 | 4.408E+02 | 4.515E+02 | EM |
| | Average | 1.981E+02 | 2.541E+02 | 2.410E+02 | 2.610E+02 | 2.890E+02 | 3.060E+02 | ECA* |
| G2-16-100 | Best | 1.213E+02 | 2.664E+02 | 2.375E+02 | 1.298E+02 | 1.963E+02 | 1.933E+02 | ECA* |
| | Worst | 2.717E+02 | 9.764E+02 | 3.725E+02 | 2.847E+02 | 2.977E+02 | 3.319E+02 | ECA* |
| | Average | 1.660E+02 | 5.410E+02 | 2.879E+02 | 1.879E+02 | 2.418E+02 | 2.012E+02 | ECA* |
| G2-1024-10 | Best | 4.493E-01 | 9.937E-01 | 7.945E-01 | 4.209E-01 | 5.996E-01 | 7.226E-01 | EM |
| | Worst | 8.730E-01 | 1.698E+00 | 1.687E+00 | 1.557E+00 | 1.417E+00 | 1.109E+00 | ECA* |
| | Average | 6.289E-01 | 1.316E+00 | 1.109E+00 | 1.100E+00 | 9.382E-01 | 8.019E-01 | ECA* |
| G2-1024-30 | Best | 8.975E-01 | 1.964E+00 | 1.662E+00 | 1.142E+00 | 1.712E+00 | 9.714E-01 | ECA* |
| | Worst | 1.954E+00 | 2.852E+00 | 2.826E+00 | 2.377E+00 | 3.924E+00 | 3.197E+00 | ECA* |
| | Average | 1.210E+00 | 2.296E+00 | 2.108E+00 | 1.917E+00 | 1.402E+00 | 1.508E+00 | ECA* |
| G2-1024-60 | Best | 2.685E+00 | 5.451E+00 | 3.580E+00 | 3.268E+00 | 3.521E+00 | 2.509E+00 | GENCLUST++ |
| | Worst | 5.155E+00 | 7.287E+00 | 7.268E+00 | 6.325E+00 | 6.524E+00 | 5.927E+00 | ECA* |
| | Average | 3.883E+00 | 6.158E+00 | 5.819E-01 | 5.186E-01 | 4.296E-01 | 3.682E+00 | GENCLUST++ |
| G2-1024-80 | Best | 2.493E+00 | 4.711E+00 | 4.841E+00 | 3.505E+00 | 2.842E+00 | 4.459E+00 | ECA* |
| | Worst | 5.913E+00 | 7.133E+00 | 8.227E+00 | 6.124E+00 | 6.297E+00 | 5.075E+00 | ECA* |
| | Average | 3.791E+00 | 5.712E+00 | 5.705E-01 | 4.392E-01 | 4.160E+00 | 4.492E-01 | ECA* |
| G2-1024-100 | Best | 3.982E+00 | 5.562E+00 | 5.305E+00 | 4.784E+00 | 4.906E+00 | 4.525E+00 | ECA* |
| | Worst | 8.359E+00 | 9.252E+00 | 8.736E+00 | 9.376E+00 | 8.223E+00 | 1.044E+01 | LVQ |
| | Average | 5.120E+00 | 7.015E+00 | 6.504E-01 | 6.613E-01 | 5.611E-01 | 6.155E-01 | ECA* |

Table 8 represents the basic statistics of the 30-solutions obtained by ECA*, KM, KM++, EM, LVQ, GENCLUST++ for execution time (Best: the best-solution, Worst: the worst solution, average: mean-solution).

Table 8: Basic statistics of the 30-solutions obtained by ECA*, KM, KM++, EM, LVQ, GENCLUST++ for execution time distance (Best: the best-solution, Worst: the worst solution, average: mean-solution)

| Datasets | Statistics | ECA* | KM | KM++ | EM | LVQ | GENCLUST++ | Winner |
|---|---|---|---|---|---|---|---|---|
| **S1** | Best | 4.2004193 | 5.5134513 | 5.3411961 | 4.4498129 | 4.8981934 | 4.3498351 | ECA* |
| | Worst | 5.1823096 | 7.1872989 | 6.8175003 | 5.3871612 | 5.2981224 | 5.8792801 | ECA* |
| | Average | 4.7901982 | 6.641405 | 6.4001498 | 5.1018735 | 4.9801157 | 5.4113987 | ECA* |
| **S2** | Best | 4.6284632 | 5.2976101 | 5.1339815 | 4.8911574 | 4.7984498 | 4.9411982 | ECA* |
| | Worst | 5.2900102 | 6.219851 | 6.1812612 | 5.5418857 | 5.4229801 | 5.7798112 | ECA* |
| | Average | 4.4009142 | 5.6618329 | 5.6891146 | 5.1872655 | 4.7812287 | 4.5411792 | ECA* |
| **S3** | Best | 4.4170244 | 4.9984761 | 5.0012701 | 5.1298043 | 5.0038192 | 4.7910191 | ECA* |
| | Worst | 5.0913972 | 5.3341802 | 5.2891713 | 5.3398101 | 5.4910914 | 4.9810043 | GENCLUST++ |
| | Average | 4.6135001 | 5.1983433 | 5.2008146 | 5.2419163 | 5.29861 | 4.8716918 | ECA* |
| **S4** | Best | 4.0876182 | 4.9273931 | 4.9853812 | 4.6109171 | 4.3918174 | 4.2097271 | ECA* |
| | Worst | 5.0552468 | 5.4801918 | 5.3712983 | 5.3309185 | 4.9800432 | 5.1150984 | LVQ |
| | Average | 4.4009821 | 5.2817161 | 5.2787173 | 4.9003873 | 4.5129827 | 4.6639823 | ECA* |
| **A1** | Best | 2.8910098 | 3.4129182 | 3.3971613 | 2.8448123 | 2.9162156 | 3.2534851 | EM |
| | Worst | 3.3590351 | 3.8987611 | 3.9123388 | 3.3800191 | 3.5660195 | 3.8122757 | ECA* |
| | Average | 2.9710084 | 3.2811757 | 3.533918 | 3.0014817 | 3.3440401 | 3.4770193 | ECA* |
| **A2** | Best | 5.2710921 | 5.5800354 | 5.4511892 | 5.4481717 | 5.3918124 | 5.6612567 | ECA* |
| | Worst | 6.3824499 | 6.8251712 | 6.9811245 | 6.4401918 | 6.7901583 | 7.1209585 | ECA* |
| | Average | 5.451176 | 6.1897163 | 6.2871655 | 5.971683 | 6.0057174 | 6.2478173 | ECA* |
| **A3** | Best | 9.0021761 | 9.6174516 | 9.7118571 | 9.4390274 | 9.1400481 | 9.4912021 | ECA* |
| | Worst | 12.0374002 | 13.0115812 | 12.541237 | 13.220973 | 11.391817 | 12.5519174 | LVQ |
| | Average | 10.1186413 | 12.1228571 | 10.4596611 | 11.0248615 | 9.9673381 | 11.4514364 | LVQ |
| **Birch1** | Best | 614.0763139 | 685.875912 | 659.289436 | 616.4739163 | 635.7442382 | 639.0509475 | ECA* |
| | Worst | 717.703542 | 725.650377 | 747.411297 | 809.2190996 | 760.9788072 | 799.2488901 | ECA* |
| | Average | 671.9864511 | 701.238791 | 721.941586 | 731.8658191 | 689.1237901 | 742.7471931 | ECA* |
| **Un-balance** | Best | 8.5611092 | 9.7398561 | 10.1026104 | 8.8535737 | 8.8253253 | 8.2295237 | GENCLUST++ |
| | Worst | 12.8915149 | 13.0892744 | 14.1022961 | 14.9140524 | 15.099935 | 13.800504 | ECA* |
| | Average | 9.6981203 | 10.9712347 | 12.8710293 | 11.8271812 | 12.4481582 | 9.8709122 | ECA* |
| **Aggregation** | Best | 0.2347718 | 0.3850435 | 0.3047076 | 0.3835216 | 0.3033671 | 0.3167278 | ECA* |
| | Worst | 0.3735991 | 0.4759776 | 0.4234431 | 0.4452728 | 0.3873649 | 0.3636616 | GENCLUST++ |
| | Average | 0.2861972 | 0.4312388 | 0.3712756 | 0.3912089 | 0.3312857 | 0.3297572 | ECA* |
| **Compound** | Best | 0.1176623 | 0.2491217 | 0.2448508 | 0.2330298 | 0.2487167 | 0.2331364 | GENCLUST++ |
| | Worst | 0.1939683 | 0.5060425 | 0.2961884 | 0.197195 | 0.2176885 | 0.2155435 | ECA* |
| | Average | 0.1461983 | 0.3257391 | 0.2601278 | 0.2007235 | 0.2307564 | 0.2254834 | ECA* |
| **Path-based** | Best | 0.2560914 | 0.3719043 | 0.326703 | 0.3116349 | 0.296982 | 0.2628713 | ECA* |
| | Worst | 0.3426792 | 0.4530889 | 0.40372 | 0.3569625 | 0.3655019 | 0.3478451 | ECA* |
| | Average | 0.2871194 | 0.4032483 | 0.3693718 | 0.3293746 | 0.3202138 | 0.2917324 | ECA* |
| **D31** | Best | 3.9761438 | 5.3121166 | 4.4743388 | 4.6925241 | 4.6744613 | 4.081507 | ECA* |
| | Worst | 5.4718194 | 6.7851171 | 6.3415327 | 6.2998281 | 5.8018699 | 6.0992068 | ECA* |
| | Average | 4.445819 | 5.8018232 | 5.5418272 | 5.1034835 | 5.1573645 | 5.4074612 | ECA* |
| **R15** | Best | 0.1487519 | 0.2259559 | 0.2222664 | 0.2015843 | 0.1847687 | 0.2236695 | ECA* |
| | Worst | 0.2852839 | 0.3766021 | 0.3655304 | 0.27897 | 0.3380696 | 0.3043674 | EM |
| | Average | 0.1909734 | 0.2957165 | 0.2815753 | 0.2493274 | 0.2525982 | 0.2609128 | ECA* |
| **Jain** | Best | 0.1448917 | 0.2602281 | 0.23896 | 0.2185464 | 0.1374285 | 0.1810869 | LVQ |
| | Worst | 0.256619 | 0.3439525 | 0.3581696 | 0.2840132 | 0.3160927 | 0.3336295 | ECA* |
| | Average | 0.1866041 | 0.2952348 | 0.2787511 | 0.2509381 | 0.17932489 | 0.2412015 | LVQ |
| **Flame** | Best | 0.1374281 | 0.2114896 | 0.1992104 | 0.2129759 | 0.2577989 | 0.2259166 | ECA* |
| | Worst | 0.2790134 | 0.3756618 | 0.3940552 | 0.3537982 | 0.3849425 | 0.3532136 | ECA* |
| | Average | 0.1908451 | 0.3218585 | 0.3004712 | 0.2712399 | 0.31324774 | 0.2537574 | ECA* |
| **Dim-32** | Best | 18.8541501 | 20.9230188 | 21.1891271 | 23.8835052 | 23.4314476 | 21.4846696 | ECA* |
| | Worst | 23.8555113 | 27.398383 | 29.0782056 | 31.5605283 | 27.2065774 | 29.6538025 | ECA* |
| | Average | 19.0186419 | 23.1617181 | 24.0918362 | 24.1812827 | 23.0071848 | 24.7817911 | ECA* |
| **Dim-64** | Best | 27.71974513 | 31.7833252 | 31.6089027 | 29.3803287 | 28.7083678 | 30.4010708 | ECA* |
| | Worst | 35.2978616 | 44.8945966 | 38.3698708 | 36.6677498 | 37.6065225 | 34.0354705 | GENCLUST++ |
| | Average | 29.9186442 | 38.9186301 | 36.0183719 | 32.1911754 | 31.8921765 | 32.01176485 | ECA* |
| **Dim-128** | Best | 82.9173461 | 84.0119322 | 85.2979698 | 85.1419189 | 87.7661945 | 83.903881 | ECA* |
| | Worst | 100.0269856 | 138.870439 | 133.272018 | 115.8113638 | 112.1446982 | 123.3893997 | ECA* |
| | Average | 89.0864163 | 97.185827 | 102.919237 | 101.0083712 | 106.1012848 | 99.6601973 | ECA* |
| **Dim-256** | Best | 178.0183741 | 191.020844 | 196.973321 | 190.4113604 | 195.1269171 | 188.4619614 | ECA* |
| | Worst | 201.2623284 | 297.65616 | 285.32109 | 248.4952963 | 216.7306471 | 214.5037416 | ECA* |
| | Average | 188.0927461 | 243.817216 | 239.771936 | 234.8167815 | 207.9187614 | 209.0018474 | ECA* |
| **Dim-512** | Best | 613.0973515 | 624.894304 | 625.50233 | 622.7406627 | 630.8232438 | 625.9239779 | ECA* |
| | Worst | 756.4176958 | 825.956128 | 823.38043 | 749.930287 | 844.2028908 | 808.9908784 | EM |
| | Average | 679.1870274 | 744.213597 | 737.017472 | 684.8176483 | 709.2148592 | 687.9873214 | ECA* |
| **Dim-1024** | Best | 709.9846101 | 773.753096 | 751.459906 | 773.3806316 | 707.9187644 | 794.1639034 | LVQ |
| | Worst | 793.7633531 | 866.284242 | 894.157274 | 787.6041536 | 815.1965552 | 854.5364531 | ECA* |
| | Average | 723.097517 | 821.876193 | 837.183762 | 746.9128764 | 753.1782762 | 809.7128653 | ECA* |
| **G2-16-10** | Best | 8.1905715 | 13.0103499 | 12.8459399 | 9.1072108 | 12.9035881 | 9.0601562 | ECA* |
| | Worst | 14.1116419 | 18.217718 | 16.3363608 | 14.0552602 | 17.4916213 | 17.9359611 | EM |
| | Average | 10.8957012 | 15.8370174 | 14.7123985 | 12.2137913 | 13.4812763 | 13.8712345 | ECA* |
| **G2-16-30** | Best | 28.9857144 | 30.9742344 | 33.5562148 | 30.2984659 | 29.5488708 | 35.734236 | ECA* |
| | Worst | 39.2827228 | 45.7586831 | 44.8095406 | 42.4028462 | 34.9099733 | 40.1282366 | LVQ |
| | Average | 32.7109165 | 38.1245972 | 37.8123743 | 34.97126213 | 32.5481237 | 36.5123874 | LVQ |
| **G2-16-60** | Best | 10.1187465 | 12.8735079 | 12.1022897 | 12.6700901 | 14.8416761 | 13.0443675 | ECA* |
| | Worst | 20.9588633 | 30.7094217 | 24.6495931 | 22.2521713 | 21.9442466 | 21.7514665 | ECA* |
| | Average | 14.0117658 | 23.1257655 | 20.5123759 | 19.6179549 | 16.7371236 | 15.7812375 | ECA* |
| **G2-16-80** | Best | 11.8563301 | 15.1590662 | 14.318351 | 16.0617645 | 16.3703685 | 16.2744314 | ECA* |
| | Worst | 20.9972697 | 31.7075329 | 30.2098935 | 21.0721911 | 21.5051813 | 22.3043732 | ECA* |
| | Average | 13.6691002 | 23.7123681 | 22.7761234 | 19.8123753 | 18.2358576 | 19.0047512 | ECA* |
| **G2-16-100** | Best | 9.1874099 | 18.9003087 | 10.0070281 | 12.6981907 | 11.6487225 | 12.93888 | ECA* |
| | Worst | 22.9424668 | 29.499466 | 23.940273 | 24.3245824 | 21.0530768 | 21.6797351 | GENCLUST++ |
| | Average | 13.0097656 | 16.9123575 | 15.7561236 | 14.9865142 | 13.9812354 | 14.8123708 | ECA* |
| **G2-1024-10** | Best | 681.8361047 | 716.776281 | 765.612157 | 814.4306876 | 736.8568727 | 800.2418677 | ECA* |
| | Worst | 854.2274421 | 988.558823 | 884.211474 | 977.2919698 | 862.4053762 | 861.4343435 | ECA* |
| | Average | 751.0198377 | 861.152349 | 811.51325 | 866.8457534 | 749.5234875 | 819.572595 | ECA* |
| **G2-1024-30** | Best | 709.2759118 | 804.02854 | 827.467382 | 710.3761094 | 843.7220831 | 723.5007255 | ECA* |
| | Worst | 948.009874 | 1012.04797 | 1061.07215 | 1024.647764 | 951.7828073 | 977.9404516 | ECA* |
| | Average | 801.7701924 | 913.958754 | 911.752372 | 798.1575699 | 877.8575641 | 789.7573112 | ECA* |
| G2-1024-60 | Best | 677.1095612 | 761.193122 | 725.567747 | 756.7170123 | 675.4049715 | 729.1513535 | LVQ |
| | Worst | 794.3868164 | 816.263899 | 814.850916 | 820.7399589 | 841.1407613 | 815.1614226 | ECA* |
| | Average | 722.0119581 | 797.986412 | 772.976421 | 802.0754631 | 754.0866467 | 735.0753278 | ECA* |
| G2-1024-80 | Best | 702.761094 | 858.964013 | 820.349344 | 764.4128872 | 703.1618522 | 745.3378721 | ECA* |
| | Worst | 861.9045881 | 983.624385 | 928.208988 | 896.7245349 | 895.6864751 | 851.9503597 | GENCLUST++ |
| | Average | 743.9188276 | 874.123858 | 856.912738 | 822.0957512 | 755.8571653 | 787.5756475 | ECA* |
| G2-1024-100 | Best | 711.0974613 | 804.372091 | 802.920492 | 741.2742261 | 765.2488254 | 824.9821503 | ECA* |
| | Worst | 831.3977194 | 973.942471 | 914.761546 | 962.1694446 | 850.6979976 | 846.478115 | ECA* |
| | Average | 778.1827655 | 834.572357 | 830.571263 | 829.5823742 | 791.5723457 | 829.5723451 | ECA* |

In brief, Table 9 summarises the results of Table 6, 7, and 8. It presents statistical comparison results using basic statistic values (Best: the best-solution, Worst: the worst solution, Average: mean-solution) obtained through 30 runs of ECA* and the comparison algorithms to solve the 31 dataset problems. These results show that ECA* is statistically performing more successful than all of the counterpart algorithms, with a statistical values (Best/ Worst/ Average) 28/ 21/ 31, 22/ 26/ 32, 26/ 21/ 31 for intraCluster, interCluster, execution time respectively.

Table 9: Performance of clustering results (intraCluster, interCluster, and execution time) of 31 datasets using ECA* compared to its competitive algorithms

| Performance parameters/ results | ECA* | KM | KM++ | EM | LVQ | GENCLUST++ |
|---|---|---|---|---|---|---|
| | Best/ Worst/ Average | Best/ Worst/ Average | Best/ Worst/ Average | Best/ Worst/ Average | Best/ Worst/ Average | Best/ Worst/ Average |
| IntraCluster | 28/ 21/ 31 | 0/ 0/ 0 | 0/ 0/ 0 | 0/ 3/ 0 | 2/ 4/ 1 | 2/ 4/ 0 |
| InterCluster | 22/ 26/ 32 | 0/ 0/ 0 | 0/ 0/ 0 | 3/ 2/ 0 | 3/ 2/ 0 | 4/ 2/ 0 |
| Execution time | 26/ 21/ 31 | 0/ 0/ 0 | 0/ 0/ 0 | 1/ 3/ 0 | 3/ 3/ 0 | 2/ 5/ 0 |

## 5.3. Performance ranking framework

In the previous section, we have studied the performance of ECA* compared to its counterpart algorithms using 32 benchmark datasets. In this section, we empirically measure how much the performance of these techniques depending on five factors: (i) Cluster overlap. (ii) The number of clusters. (ii) Cluster dimensionality. (iv) Well-structured clusters (structure). (v) Cluster shape. The measure of ECA*'s overall performance with its competitive techniques in terms of the above factors is presented in the form of framework. These factors have been also studied by [2] for evaluating different clustering algorithms. Figure 6 presents the overall average performance rank of ECA* with its counterpart algorithms. In this figure, we rank these techniques based on the above factors using internal measures and external measures. For both internal and external measures, rank 1 represents the best performing algorithm to a current factor, whereas rank 5 refers to the worst-performing algorithms to a current factor.

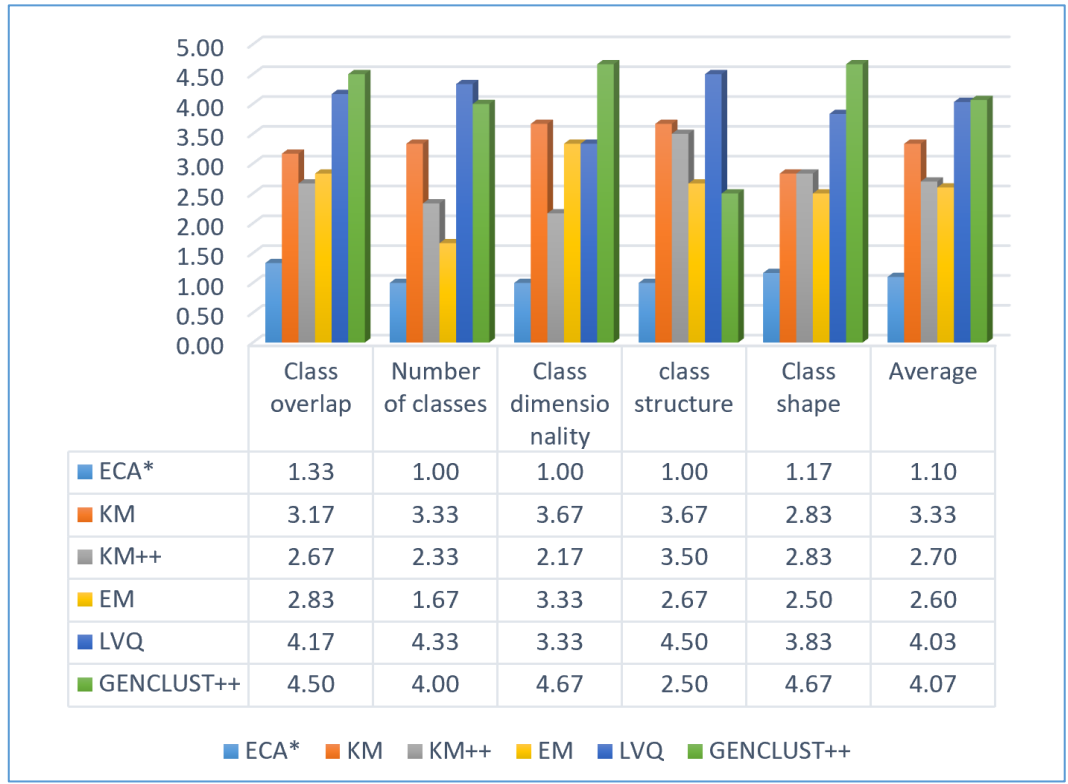

**Figure 5: Overall average performance rank of ECA\* compared to its counterpart algorithms for 32 datasets according to five dataset features**

The results show that ECA* outperforms its competitive techniques in all the mentioned factors. After ECA*, three different techniques have scored the second successful technique according to the five mentioned factors: (1) EM is considered as a useful technique to solve the dataset problems that have cluster overlap and cluster shape. (2) KM++ works well for the datasets that have a number of clusters, and cluster dimensionality. (3) GENCLUST++ does well for the datasets that have good cluster structure. Figure 6 depicts a performance ranking framework of ECA* compared to its counterpart algorithms for 31 multiple-featured datasets according to these features (cluster overlap, cluster number, cluster dimensionality, cluster structure, and cluster shape). In this figure, the dark colours show that the algorithm is well performed (ranked as 1) for a specified dataset features, whereas the light colours indicate that the technique is poorly performed (ranked as 6) for a specified dataset property. The colours in between dark and light are ranked as 2, 3, 4, and 5, respectively. Overall, the rank of the algorithm for all the five dataset features is ECA*, EM, KM++, KM, LVQ, and GENCLUST++.

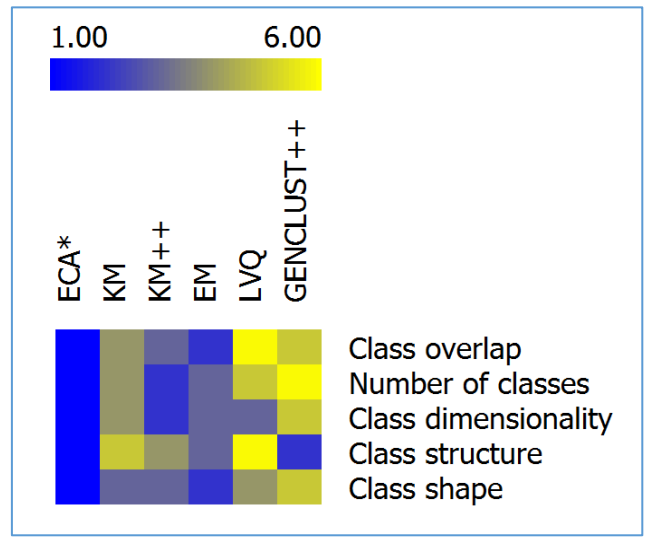

**Figure 6: A performance ranking framework of ECA* compared to its counterpart algorithms for 32 multiple-featured datasets.**

## 5.4. Success and limitations of ECA*

The experimental results are presented to discuss the strengths and weaknesses of the ECA*. On the one hand, the ECA* has several strengths compared with its counterpart algorithms:

1. The ECA* is useful for understanding clustering results for diverse and multiple-featured benchmarking problems. In other words, the dimensionality of problems and different dataset features, such as cluster overlap, the number of clusters, cluster dimensionality, the cluster structure, and the cluster shape, do not affect the effectiveness of the ECA*.

2. The proposed ECA* method increases performance due to its mut-over strategy. Owing to the use of this strategy, the results show that this method can effectively search for possible solutions and enhance the effectiveness of cluster analysis. This strategy can generate good cluster centroids based on the intraCluster and interCluster distances and serve as a boundary control mechanism beyond the search space limitations to promise population diversity and ensure effective searches for producing good clustering results and cluster centroids. Hence, the use of mut-over in the ECA* can produce more reliable results than those obtained by its competitive clustering algorithms.

3. The clustering method is an unsupervised learning algorithm that seems useful in identifying excellent results in the absence of pre-defined and prior knowledge. Therefore, combining different optimisation techniques in the ECA* by considering the boundary control mechanism concerning the objective functions can overcome the instability inherent in clustering algorithm problems. Accordingly, this prevents the algorithm from becoming trapped in local and global optima.

4. Another remarkable achievement of the ECA* is the use of a dynamic clustering strategy to find the right number of clusters. In this strategy, the ECA* removes empty clusters, merging dynamic clusters that are less dense with their neighbouring clusters and merging two clusters based on their distance measures.

5. The use of statistical techniques in the ECA* helps to adjust the clustering centroids towards their objective function and results in excellent clustering centroids accordingly.

On the other hand, the results of the experiments conducted in this study recognise some weaknesses of the ECA*:

1. Initialising the optimal value for the pre-defined variables in the ECA*, such as the number of social class ranks and the cluster density threshold, is a difficult task. Choosing an optimal number of social class ranks may result in a failure to determine the optimal number of clusters. In addition, the ECA* lacks in choosing the right cluster density threshold, which can be suitable for complex and multiple-featured problems. The cluster density threshold can be different based on the type of benchmarking problem. For instance, a cluster density threshold of 0.001 might be optimum for one type of dataset, while it might be non-optimal for another type of dataset. Therefore, the determination of the cluster density threshold should be adopted based on the scale and features of the dataset.

2. Choosing a different number of social class ranks might affect how the cluster threshold density is defined. A small number of social class ranks can result in a low number of clusters and a high cluster density threshold, whereas a large number of social class ranks can result in a high number of clusters and a low cluster density threshold. Since social class ranks and the cluster density threshold are pre-defined values, balancing between these two parameters can be a challenging task.

3. The cluster merging process is not always accurate. Two clusters can be merged based on their average intraCluster and the minimum distance between them. Sometimes, the number of generated clusters and the cohort of benchmark datasets might affect the accuracy of the merging process.

4. Since ECA* is considered as an integrated multi-disciplinary algorithm, this may sometimes bring a reluctant conclusion that multi-disciplinary may not facilitate the use of ECA* by single discipline scholars and professionals.

## 6. Conclusion

In this study, we have proposed a new evolutionary clustering algorithm based on social class ranking, quartiles, percentiles, evolutionary algorithm operators (BSA and LFO), and the Euclidean distance in the K-means algorithm. We have also examined the performance of the ECA* based on 32 heterogeneous and multiple-featured datasets in comparison with five other classic and state-of-the-art algorithms with the use of internal and external validation criteria and basic statistical performance measures. The evidence of the experimental studies indicates two primary results. (i) None of its counterpart algorithms is superior to the ECA* with regard to their capacity to recognise the right clusters of functionally related observations in the given 32 benchmarking datasets. (ii) The proposed performance framework indicates that the ECA* is a more competitive algorithm than all of the other techniques, and it is well suited for complex and multiple-featured benchmarking problems. Based on the strengths and limitations of the ECA*, we suggest several avenues for future studies. Future research can optimise the cluster results by combining various problem sources, such as numerical and categorical datasets, and real-world problems. Additionally, it can apply the ECA* under the assumption of no prior knowledge, for example, regarding the number of social classes and cluster density threshold since only one data source is used for the fitness operation. Accordingly, we intend to develop a proper fitness procedure to study combined multi-source datasets and a new analysis process. Finally, exploiting and adapting the ECA* for real-world applications is a vital possibility for future research, such as ontology learning in the Semantic Web, practical application problems [38], technical application problems[39], and e-government services [40].

**Acknowledgements**

The authors would like to thank the referees for their remarkable suggestions. This paper's technical content has significantly improved based on their suggestions. Meanwhile, the authors wish to express a sincere thanks to Kurdistan Institution for Strategic Studies and Scientific Research and the University of Kurdistan Hewler for providing facilities and continuous support in conducting this study.

**Compliance with Ethical Standards**

**Conflict of interest:** The authors declare that they have no conflict of interest.

**Funding:** Funding information is not applicable / No funding was received.